\documentclass[final,5p,times,twocolumn]{elsarticle}

\pdfoutput=1
\DeclareUnicodeCharacter{2217}{\'{*}}
\usepackage{soul}
\soulregister\cite7 
\soulregister\citep7 
\soulregister\citet7 
\soulregister\ref7 
\soulregister\pageref7 
\usepackage{natbib}
\usepackage{colortbl}
\usepackage{amsmath,amsfonts}
\usepackage{caption}
\usepackage{subcaption}
\usepackage{hyperref}
\hypersetup{hypertex=true,
            colorlinks=true,
            linkcolor=red,
            anchorcolor=blue,
            citecolor=green}
\usepackage{array}
\usepackage{arydshln}
\usepackage{textcomp}
\usepackage{stfloats}
\usepackage{url}
\usepackage{verbatim}
\usepackage{graphicx}
\usepackage{cite}
\usepackage{bbding}
\usepackage{multirow}
\usepackage{makecell}
\usepackage{soul}
\soulregister\cite7 
\soulregister\citep7 
\soulregister\citet7 
\soulregister\ref7 
\soulregister\pageref7 
\usepackage{xcolor}
\usepackage{setspace}
\usepackage{nicematrix}
\usepackage{todonotes}
\usepackage{tabularx}
\usepackage{booktabs}
\usepackage{tabularray}
\usepackage{enumitem}  
\usepackage{sidecap}
\usepackage{tikz}

\usepackage{mathrsfs}
\usepackage[noend]{algpseudocode}
\usepackage{multirow}

\usepackage{algorithmicx,algorithm}
\usepackage{bbm}
\usepackage{graphicx}
\usepackage{amsmath}
\usepackage{amssymb}
\usepackage{multirow}
\usepackage{helvet}
\usepackage{courier}
\usepackage{float}
\usepackage{color,soul,xcolor}

\usepackage{amsfonts}
\usepackage{textcomp}

\usepackage{times}
\usepackage{epsfig}

\usepackage{array}
\usepackage{textcomp}
\usepackage{stfloats}
\usepackage{url}
\usepackage{verbatim}
\usepackage{arydshln}
\usepackage{bbding}
\usepackage{wrapfig}
\usepackage{caption}
\usepackage{booktabs}
\usepackage{colortbl}
\usepackage{newfloat}
\usepackage{listings}
\usepackage{bm}

\usepackage{makecell}
\usepackage{pifont}
\usepackage{stmaryrd}
\usepackage{tabularx}

\usepackage{tikz}
\usepackage{tabularray}
\definecolor{mycolor1}{RGB}{237,125,49}
\definecolor{MineShaft}{rgb}{0.2,0.2,0.2}
\newcommand{\firstcolor}[1]{\colorbox{red!15}{#1}}

\newcommand{\mycolor}[2][1]{
    \textcolor[rgb]{#1,0,0}{#2}
}

\journal{Expert Systems with Applications}

\begin{document}
\begin{sloppypar}
\begin{frontmatter}

\title{FDBPL: Faster  Distillation-Based Prompt Learning for Region-Aware Vision-Language Models Adaptation}


\author[bupt_address]{Zherui {Zhang}\fnref{fn1}}
\author[address0]{Jiaxin {Wu}\fnref{fn1}}
\author[4,5]{Changwei Wang}
\author[2]{Rongtao Xu}
\author[bupt_address]{\\Longzhao Huang} 
\author[bupt_address]{Wenhao Xu}
\author[bupt_address]{Wenbo Xu}
\author[bupt_address]{Li Guo}
\author[bupt_address]{Shibiao Xu\corref{cor1}}
\cortext[cor1]{Corresponding Author
  Email: shibiaoxu@bupt.edu.cn}
\fntext[fn1]{Zherui {Zhang} and Jiaxin {Wu} contribute equally}

\address[bupt_address]{School of Artificial Intelligence, Beijing University of Posts and Telecommunications, China}

\address[address0]{School of National Elite Institute of Engineering, Chongqing University, Chongqing, China}

\address[2]{The State Key Laboratory of Multimodal Artificial Intelligence Systems, Institute of Automation, Chinese Academy of Sciences, Beijing, China}
\address[4]{The Key Laboratory of Computing Power Network and Information Security, Ministry of Education, Shandong Computer Science Center, Qilu University of Technology, Jinan, China.}
\address[5]{Shandong Provincial Key Laboratory of Computing Power Internet and Service Computing, Shandong Fundamental Research Center for Computer Science, Jinan, China.
\vspace{-1cm}
}

\begin{abstract}
Prompt learning as a parameter-efficient method that has been widely adopted to adapt Vision-Language Models (VLMs) to downstream tasks. While hard-prompt design requires domain expertise and iterative optimization, soft-prompt methods rely heavily on task-specific hard labels, limiting their generalization to unseen categories. Recent popular distillation-based prompt learning methods improve generalization by exploiting larger teacher VLMs and unsupervised knowledge transfer, yet their repetitive teacher model online inference sacrifices the inherent training efficiency advantage of prompt learning. 
In this paper, we propose {{\large {\textbf{F}}}}aster {{\large {\textbf{D}}}}istillation-{{\large {\textbf{B}}}}ased {{\large {\textbf{P}}}}rompt {{\large {\textbf{L}}}}earning (\textbf{FDBPL}), which addresses these issues by sharing soft supervision contexts across multiple training stages and implementing accelerated I/O.  
Furthermore, FDBPL introduces a region-aware prompt learning paradigm with dual positive-negative prompt spaces to fully exploit randomly cropped regions that containing multi-level information. We propose a positive-negative space mutual learning mechanism based on similarity-difference learning, enabling student CLIP models to recognize correct semantics while learning to reject weakly related concepts, thereby improving zero-shot performance.
Unlike existing distillation-based prompt learning methods that sacrifice parameter efficiency for generalization, FDBPL maintains dual advantages of parameter efficiency and strong downstream generalization. Comprehensive evaluations across 11 datasets demonstrate superior performance in base-to-new generalization, cross-dataset transfer, and robustness tests, achieving $2.2\times$ faster training speed.
\end{abstract}
\begin{keyword}
Vision-Language Models, Prompt Learning, Knowledge Distillation.
\end{keyword}

\end{frontmatter}


\section{Introduction}
\label{Introduction}
With the rapid development of deep learning technology, traditional models that rely only on visual features for precise recognition are no longer sufficient to meet the diverse task requirements found in open-world scenarios.
Therefore, recent large-scale pre-trained Vision-Language Models (VLMs), such as CLIP~\citep{radford2021learning}, DALL-E~\citep{ge2022dall}, and Stable Diffusion~\citep{rombach2021highresolution}, allow them to flexibly handle various downstream tasks~\citep{Zhang2024MIMHDMS}, including image classification~\citep{sun2025reusable, ma2024coordinate, lu2024disentangling}, object detection~\citep{yu2025attention, su2025balanced, tan2024efficient}, semantic segmentation~\citep{chen2023consistency, liu2025awardistill, xiao2025domain, guo2025multilevel}, Embodied Artificial Intelligence~\citep{Xu2024DefFusionDM,Wang2024ExploringID,ren2024infiniteworld,xu2025a0,liang2025structured} and visual question answering~\citep{xu2025diff, han2025lrcn}. Specifically, CLIP employs a dual-tower architecture, comprising an image encoder and a text encoder, which combines natural language processing with computer vision and has completed extensive self-supervised pre-training on the vast amount of image-text pairs.

\begin{figure}[!t]
\centering
\includegraphics[width=0.95\linewidth]{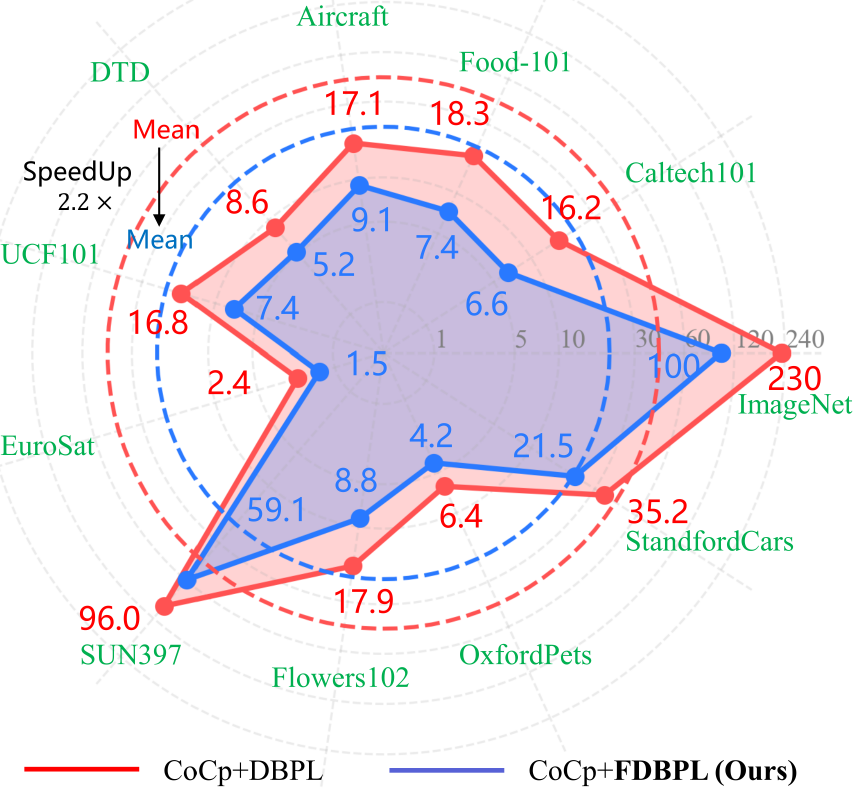}
\caption{\textbf{Training Efficiency Advantage.}
The results are measured in minutes, and our method demonstrates significant improvements across all 11 datasets, as well as in average training time.
}
\label{fig:f1}
\end{figure}

\begin{figure*}[!t]
\centering
\includegraphics[width=\linewidth]{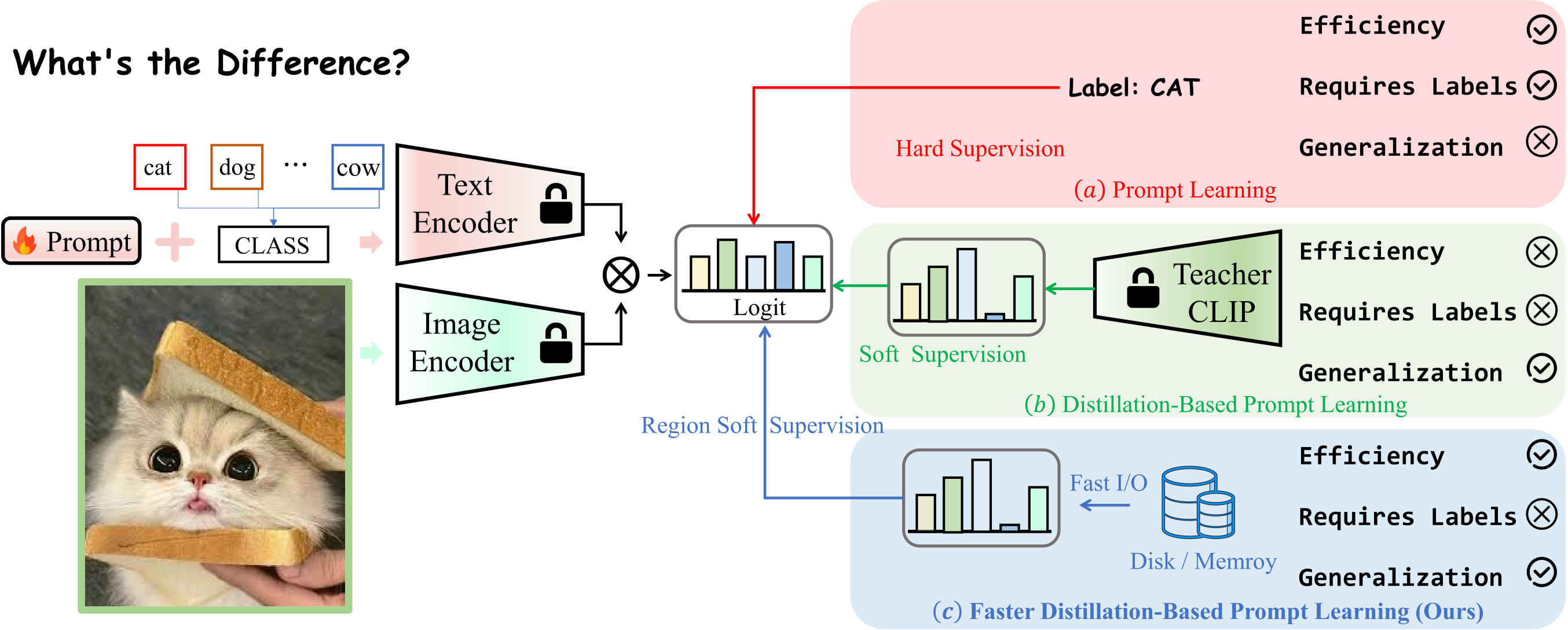}
\caption{\textbf{From PL to FDBPL~(Ours)}.
Prompt learning (PL) adapts CLIP to downstream tasks via learnable parameters. We compare three methods:
\textbf{\textit{(a)}} Native PL employs dataset hard labels, offering fast training but suffering from overfitting on seen classes, which degrades zero-shot performance on unseen classes.
\textbf{\textit{(b)}} Distillation-Based PL uses a teacher CLIP network to transfer generalization knowledge without specific labels, often on unlabeled data. However, the online inference required for soft label generation reduces its training efficiency.
\textbf{\textit{(c)}} Our proposed Faster Distillation-Based PL~(FDBPL) achieves both high training efficiency and strong zero-shot generalization by pre-storing shareable soft labels across epochs and employing fast I/O.
}

\label{fig:f2}
\centering
\end{figure*}

VLMs like CLIP demonstrate remarkable zero-shot abilities, but require adaptation for optimal performance on downstream tasks. Among various fine-tuning explorations, prompt learning has emerged as a parameter-efficient technique that preserves CLIP's frozen dual encoder architecture and adjusts only the prompt behavior, generally falling into two categories: hard-prompt learning and soft-prompt learning.
Standard hard-prompt learning employs fixed templates (e.g., \textit{``a photo of a [class name]"}) to guide the text encoder in generating class-specific text features, where visual-textual similarity determines final predictions. 
However, hard-prompt learning suffers from critical limitations: designing optimal hard-prompts requires extensive expert knowledge or trial-and-error adjustments, making it an art rather than a science~\citep{zhang2024amend}. 
In contrast, learnable soft-prompt learning addresses this challenge by exploiting few-shot training data to optimize class-wise context vectors through contrastive alignment between visual and textual features in a shared embedding space, demonstrating superior performance over manual template engineering.
ProGrad~\citep{grad} and KgCoOp~\citep{kgcoop23} reveal critical limitations in both paradigms: soft-prompts achieve strong in-domain performance in base classes, tend to overfit training data and show a weak generalization to unseen classes\citep{mistretta2024improving,li2024promptkd}, and hard-prompts maintain better zero-shot performance through their explicit semantic grounding. 
{{\textbf{LAMM}} \citep{gao2024lamm} introduces an innovative label alignment that dynamically adjusts the category embeddings of downstream datasets through end-to-end training.
{\textbf{GIST}} \citep{ruan2024gist} addresses the behavioral gap between trainable parameters and downstream task knowledge. }
The above observation raises a critical challenge:  how can we effectively exploit the advantages of both hard and soft prompt learning?
Consequently, another branch of prompt learning explores the extraction of knowledge from more powerful VLMs as teachers, thereby reducing the need for labeled data for target downstream tasks. Figure~\ref{fig:f2}~\textbf{(b)} demonstrates how prompt learning and knowledge distillation are combined to improve the zero-shot adaptation performance of the student CLIP in downstream tasks.

In this paper, we follow the design paradigm of Distillation-Based Prompt Learning, but we emphasize that the inherent efficiency advantages of native prompt learning are impaired by the introduction of knowledge distillation. 
As depicted in Figure~\ref{fig:f2}~\textbf{(b)}, knowledge distillation, acting as an additional supervisory objective for the student CLIP, introduces a non-negligible efficiency degradation (i.e., longer training time per epoch), due to obtaining such supervisory signals inevitably necessitates online inference from the larger teacher VLMs. In other words, a considerable portion of computational resources is dedicated to feeding training data into the large teacher network to generate supervisory information instead of updating the student CLIP's prompts, and this supervisory information is repeatedly computed across different training epochs.
Therefore, the central problem addressed in this paper is how to leverage knowledge distillation to improve the performance of soft-prompt learning on zero-shot downstream tasks while preserving the native training efficiency advantages.

\begin{figure}[!t]
\centering
\includegraphics[width=\linewidth]{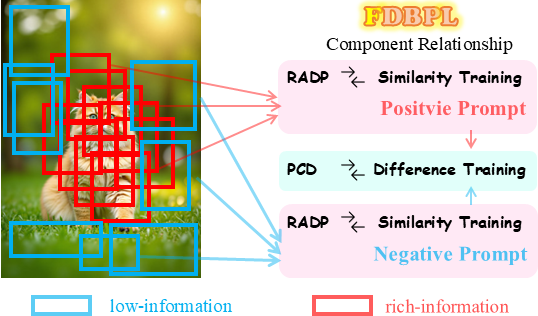}
\caption{\textbf{Relationships between Components within FDBPL.}
The training paradigm based on random region images inevitably include regions with insufficient information content (blue areas) compared to well-defined regions of interest (ROI) under sharp distributions (red areas). To address this challenge, we implement RADP (Region-Adaptive Dual Prompt) learning, which uses positive prompts for high-information regions and negative prompts for low-information regions, dual-prompt method enables independent similarity training for distinct region types. 
Furthermore, we introduce PCD (Positive-Contrastive Discrimination) learning, which uses positive-negative space contrastive analysis to capture both intra-class and inter-class latent relationships, thereby benefiting complex recognition scenarios.
}
\label{fig:f3}
\end{figure}

We propose the {{\large {\textbf{\textcolor{red}{F}}}}}aster {{\large {\textbf{D}}}}istillation-{{\large {\textbf{B}}}}ased {{\large {\textbf{P}}}}rompt {{\large {\textbf{L}}}}earning (\textbf{\textcolor{red}{F}DBPL}) framework, whose core idea adopts a space-for-time strategy to fast I/O share computations of teacher supervision signals across different training iterations, as shown in Figure~\ref{fig:f2}~\textbf{(c)}.
The proposed FDBPL comprises two key components: a Region-Aware Dual-Prompt~(RADP) learning strategy and a Prompt-Cascaded Difference~(PCD) module. 
Specifically, we randomly crop regions of interest (ROI) from the given image, store the absolute spatial position information of the cropped ROIs, the type of image transformation, and the corresponding teacher supervision information. However, due to the randomness of cropping, low-information regions inevitably emerge, which may lack the necessary semantic content. Unlike the rough discard operation~\citep{shen2022fast,shen2023ferkd}, in the proposed RADP learning, in addition to learning positive semantics through the positive prompt, we introduce learnable negative prompts that establish a greater similarity to cropped regions without explicit semantic information, teaching the student CLIP to say "No".
On the other hand, the PCD learning analyze first-order and second-order difference similarities between image and text features, capturing potential 1) intra-class differences between the presence and absence of semantics and 2) inter-class structural semantic differences, respectively, in the difference space, which achieve distinguish subtle semantic variations, particularly for categories that are semantically similar but distinct.
Figure~\ref{fig:f3} shows the relationship between the RADP and PCD components within FDBPL.
Our contributions are as follows:
\begin{itemize}
    \item  \textbf{Faster Distillation-based Prompt Learning:} We observe the limitations of existing distillation-based prompt learning methods that require repeated inference from teacher networks across multiple iterations to obtain soft supervision. 
    Our FDBPL reduces these limitations by implementing fast I/O operations to shared soft-contexts and effectively exploits unsupervised region images to improve zero-shot performance on novel classes.
    \item \textbf{Prompot Similarity-Difference Training:} We introduce dual-prompt spaces (positive and negative) that employ Region-Aware Dual Prompt~(RADP) learning to enable the student CLIP to master the correct semantic categories and learn to say ``No" to uncertain input. 
    We further incorporate Prompt-Cascaded Difference~(PCD) Learning, which facilitates mutual learning between positive and negative spaces to capture inherent intra-class and inter-class relationships, thus improving zero-shot performance in complex scenarios.
    \item {\textbf{Dual Benefits in Accuracy and Training Efficiency:}} 
    We conduct extensive experiments across 11 downstream datasets, including base-to-new transfer, cross-dataset evaluation and robustness testing, which demonstrate the significant advantages of our proposed FDBPL in zero-shot recognition tasks. Furthermore, compared to existing state-of-the-art distillation-based prompt learning methods, FDBPL achieves an average $\mathbf{2.2} \times$ acceleration in training speed, as shown in Figure~\ref{fig:f1}.
\end{itemize}

\noindent
{
The remaining structure of our paper is organized as follows:}
\begin{itemize}
    \item {\textbf{Section~\ref{relat} (Related Work)} contextualizes our Faster Distillation-Based Prompt Learning (FDBPL) framework within recent Knowledge Distillation~(Section~\ref{KD}) and Prompt Learning~(Section~\ref{pl}) advances, highlighting our novel contributions.}
    
    \item {\textbf{Section~\ref{method} (Method)} presents prompt learning $\rightarrow$ Distillation-Based Prompt Learning $\rightarrow$ \textbf{FDBPL~(Ours)}. This section details the Region-Aware Dual Prompt~(RADP, Section~\ref{RADP}) learning and Prompt-Cascaded Difference~(PCD, Section~\ref{PCD}) Learning essential to our method.}
    
    \item {\textbf{Section~\ref{Experiments} (Experiments)} evaluates FDBPL through zero-shot performance comparisons and training efficiency against state-of-the-art methods across 11 datasets, supplemented by ablation studies validating each component's contribution~(Section~\ref{Ablation Study}).}
    
    \item {\textbf{Section~\ref{sec:con} (Conclusion)} summarizes our contributions and outlines future research directions.}
\end{itemize}

\section{Related Works}
\label{relat}

\subsection{Knowledge Distillation}
\label{KD}
Knowledge distillation (KD) effectively transfers knowledge from larger pre-trained teacher networks to compact student networks, achieving competitive performance with reduced computational costs across various domains including image classification~\citep{Zhang2025CAEDFKDBT,10980222,Zhang2024MIMHDMS}, semantic segmentation~\citep{Wang2023CNDescCN,Wang2025FocusOL,}, and object detection~\citep{Xu2025TokenMT}. Recent studies explore KD applications in vision-language models (VLMs), particularly CLIP-based frameworks. SF-CLIP~\citep{sameni2024building} introduces feature mask distillation through mask reconstruction tasks, while MobileCLIP~\citep{vasu2024mobileclip} improves multi-teacher contrastive knowledge distillation with synthetic captions and data augmentation. TinyCLIP~\citep{wu2023tinyclip} proposes affinity mimicry-based weight inheritance, and CLIP-KD~\citep{Clip-kd} systematically investigates alignment objectives in KD. CLIP-CID~\citep{yang2024clip} further advances knowledge transfer through image semantic balancing and instance clustering.

Existing KD methods prioritize mainly student network compactness while overlooking distillation efficiency. FKD~\citep{shen2022fast} improves KD efficiency through random image cropping, offline alignment objective storage, and label compression. FerKD~\citep{shen2023ferkd} extends this by implementing multi-level management that discards low-information-cropped images. 
Unlike these crop-discarding attempts, our \textbf{FDBPL} framework introduces a region-aware dual-prompt learning strategy that fully exploits low-information crop regions to teach CLIP models negative understanding skills without sacrificing any cropped images.

\subsection{Prompt Learning}
\label{pl}
Vision-Language Models (VLMs), particularly CLIP, have revolutionized computer vision through dual-tower architecture design. 
Prompt learning adapts these VLMs to downstream tasks by reformulating them as ``fill-in-the-blank" problems, requiring minimal parameters while leveraging CLIP's inherent knowledge. Early hard-prompt methods rely on manually designed templates (e.g., "a photo of [CLASS]"), necessitating expert intervention and limiting flexibility.

To overcome this limitation, soft-prompt learning methods have been developed: CoOp~\citep{zhou2022learning} learns continuous context vectors that adapt CLIP's text encoder to target datasets while keeping the dual-tower architecture frozen. CoCoOp~\citep{zhou2022conditional} extends this approach by integrating visual features as implicit conditions for prompt generation. MaPLe~\citep{khattak2023maple} establishes a joint vision-language prompting framework through dual-branch architecture, while PromptSRC~\citep{khattak2023self} enhances text diversity via prompt regularization. 
{
\textbf{LAMM} \citep{gao2024lamm} dynamically adjusts the category embeddings of downstream datasets through end-to-end training, aiming to achieve a more appropriate label distribution.
\textbf{GIST} \citep{ruan2024gist} introduces a trainable token, termed the GIST token, which aggregates task-specific knowledge learned by Parameter-Efficient Fine-Tuning (PEFT) methods, thereby forming an explicit association with downstream knowledge.
}
However, current soft-prompt methods depend mainly on labeled downstream task data, limiting their generalization to unseen categories or datasets. Recent work addresses this through unlabeled knowledge transfer: UPL~\citep{huang2022unsupervised} implements label-free prompt learning, while KDPL~\citep{mistretta2024improving} and PromptKD~\citep{li2024promptkd} leverage teacher networks to transfer open-world knowledge using few-shot unlabeled dataset.

We identify a critical efficiency bottleneck in existing distillation-based prompt learning methods that requires  online repeatedly inference of teacher networks. Our FDBPL resolves this conflict through a space-time sharing strategy, demonstrating that prompt learning and knowledge distillation can collaboratively improve both training efficiency and downstream zero-shot performance, as shown in Figure~\ref{fig:f2}~\textbf{(c)}.

\begin{figure*}[!t]
\centering
\includegraphics[width=0.9\linewidth]{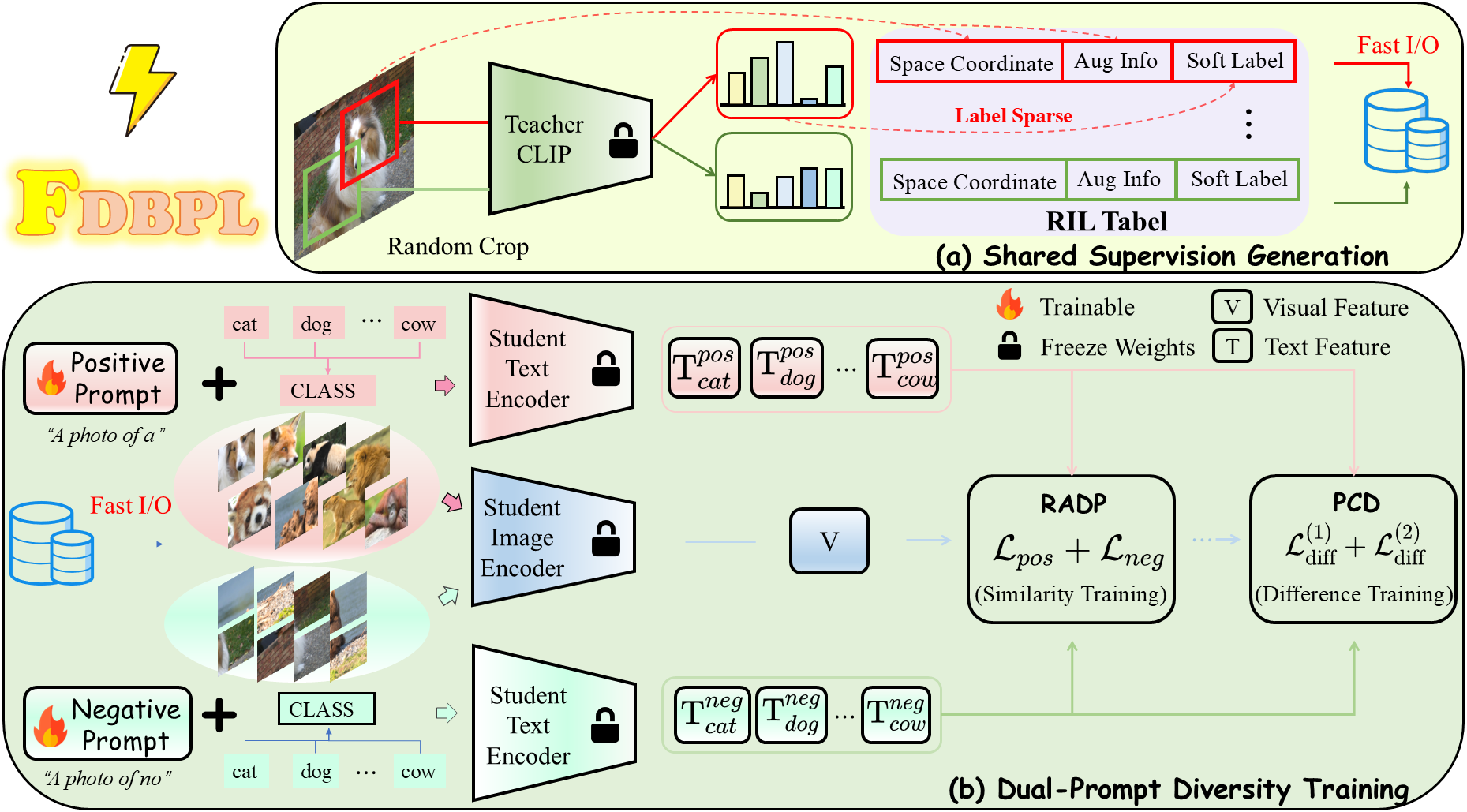}
\caption{\textbf{FDBPL Framework}.
\textbf{\textit{(a)}} To mitigate native prompt learning's strong dependency on hard-labeled downstream dataset, we introduce a larger teacher CLIP network that transfers generalized knowledge through unlabeled regional images. For efficient knowledge distillation, we pre-store spatial coordinates of randomly cropped sub-regions, data augmentation types, and teacher-generated soft labels in a Region Information Lookup (RIL) Table - a design that eliminates redundant online inference by sharing soft labels across training epochs. Label sparsification is adopted to prevent I/O bottlenecks caused by excessive soft-label storage requirements.
\textbf{\textit{(b)}} Through direct retrieval of regional images and shared supervision signals from storage devices, we develop Region-Aware Dual-Prompt Learning (RADP) with learnable positive-negative prompts that independently align information-rich and information-poor regions, dual-similarity mechanism enables the student CLIP model to simultaneously recognize correct semantic categories and reject uncertain regions.
{Specifically, within RADP, one path involves the student image encoder processing image regions that possess clear and distinguishable semantic content, as depicted in the pink input stream}~\textcolor{mycolor1}{\rule{1em}{1em}}.
{Concurrently, another path involves the student image encoder receiving image regions that lack clear semantic content, such as background areas resulting from random cropping, object edge portions, or regions which are otherwise difficult to classify, as shown in the green input stream}~\textcolor{green}{\rule{1em}{1em}}. 
Subsequently, a Prompt-Cascaded Difference~(PCD) Learning module establishes cascaded difference spaces: first-order and second-order difference spaces that respectively capture intra-class variations and inter-class relationships, thereby enhancing zero-shot recognition capabilities in complex scenarios.
}

\label{fig:f4}
\end{figure*}

\section{Method}
\label{method}

\subsection{Problem Formulation and Overview}

\vspace{2mm}
\noindent
\textbf{Native Soft-Prompt Learning.}  
Let $\mathbf{X}$ denote the image space and $\mathcal{Y} = \{l_1, \ldots, l_C\}$ the label space with $C$ classes. The CLIP dual-encoder architecture contains an image encoder $f$ and text encoder $g$ that respectively extract features $f(\mathbf{x})$ for image $\mathbf{x} \in \mathbf{X}$ and $g \circ \mathbf{V}(l_c)$ for class $l_c \in \mathcal{Y}$, where $\mathbf{V}$ represents the prompt template and $\circ$ denotes function composition.

As shown in Figure~\ref{fig:f2}~\textbf{(a)}, Following CoOp's soft-prompt learning paradigm, we define the learnable prompt for class $l_c$ as:
\begin{equation}
    \mathbf{V}(l_c) = \begin{bmatrix} \mathbf{v}_1, \mathbf{v}_2, \ldots, \mathbf{v}_L, \mathbf{w}_c \end{bmatrix}
\end{equation}
where $\{\mathbf{v}_i\}_{i=1}^L$ are learnable vectors sharing the embedding dimension (analogous to context tokens like "a photo of a"), and $\mathbf{w}_c$ is the class token for $l_c$. The predicted probability for training sample $(\mathbf{x}_i, l_i)$ is computed as:
\begin{equation}
    \begin{aligned}
        p_i = \frac{\exp\left( \cos\left( f(\mathbf{x}_i), g \circ \mathbf{V}(l_i) \right)/\tau \right)}{\sum_{l_c \in \mathcal{Y}} \exp\left( \cos\left( f(\mathbf{x}_i), g \circ \mathbf{V}(l_c) \right)/\tau \right)}
    \end{aligned}
    \label{eq:prob}
\end{equation}
where $\tau$ denotes the temperature parameter. The cross-entropy loss optimizes the prompt parameters while freezing CLIP encoders:
\begin{equation}
    \begin{aligned}
        \mathcal{L}_{\text{N}} = -\mathbb{E}_{(\mathbf{x}_i, l_i) \in (\mathbf{X},\mathcal{Y})} \log p_i
    \end{aligned}
\end{equation}

\vspace{2mm}
\noindent
\textbf{Distillation-Based Prompt Learning.}  
While native prompt learning improves downstream task performance, its reliance on labeled data degrades CLIP's generalization to unseen classes. To address this, as shown in Figure~\ref{fig:f2}~\textbf{(b)}, KDPL and PromptKD employ a larger CLIP teacher model that transfers knowledge to a smaller student CLIP on the unlabeled dataset.

\vspace{2mm}
\noindent
{\textbf{Motivation.}}
Although distillation-based prompt learning improves zero-shot recognition, the requirement for online teacher inference (through both image and text encoders) introduces substantial computational overhead that compromises the inherent efficiency advantages of native prompt learning, and training efficiency bottleneck motivates our \textbf{FDBPL} design that maintains distillation benefits while restoring adaptation efficiency.


\vspace{2mm}
\noindent
{\textbf{FDBPL Overview.}}
As illustrated in Figure~\ref{fig:f4}, FDBPL accelerates distillation-based prompt learning through two phases:
\begin{enumerate}[label=\alph*.]
    \item \textit{Shared Supervision Generation} (Figure~\ref{fig:f4}~\textbf{(a)}): We pre-compute and store teacher soft-supervision in a Region Information Lookup (RIL) Table, which records spatial coordinates of cropped regions and corresponding data augmentation types. Two label sparsity strategies (Section~\ref{Shared Supervision Generation}) prevent I/O bottlenecks caused by redundant soft labels, and this one-time process does not incur additional training overhead and supports offline updates.
    \item \textit{Dual-Prompt Diversity Training} (Figure~\ref{fig:f4}~\textbf{(b)}): The student CLIP directly fast retrieves soft-supervision from the RIL Table. We propose two complementary modules: \textbf{\textit{i)}} Region-Aware Dual-Prompt~(RADP, Section~\ref{RADP}) Learning  employs similarity-based training to teach the student say ``Yes" or ``No" to regional image; 
    \textbf{\textit{ii)}} Prompt-Cascaded Difference~(PCD, Section~\ref{PCD}) Learning employs difference-based training to facilitate the student's understanding of the latent intra-class and inter-class semantic relationships within the positive-negative prompt difference space.
\end{enumerate}

\subsection{Shared Supervision Generation}
\label{Shared Supervision Generation}
The computational burden introduced by online inference of teacher network in ``Distillation-Based Prompt Learning" compromises the inherent efficiency advantages of native ``Prompt Learning" frameworks, particularly when the same supervisory signals are re-computed redundantly across training epochs. To resolve this inefficiency, we implement a space-for-time improvement strategy that pre-computes and stores shared contextual knowledge offline.

\begin{figure*}[!t]
\centering
\includegraphics[width=0.9\linewidth]{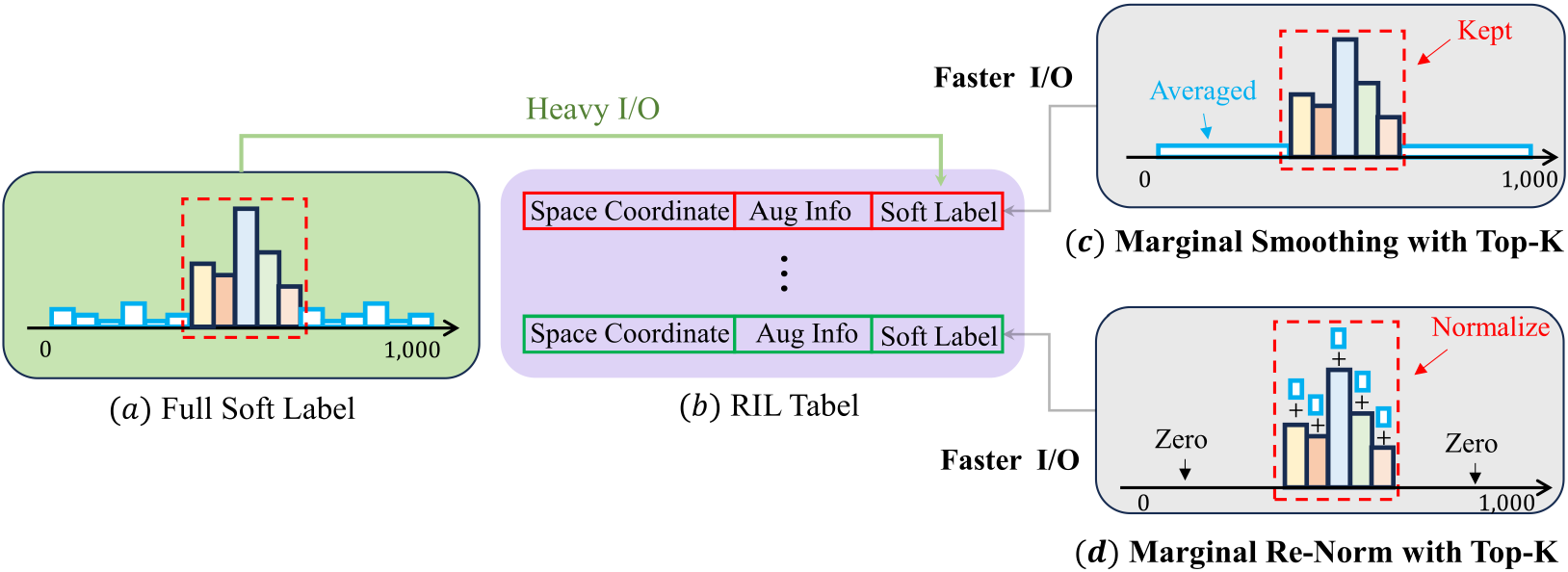}
\caption{\textbf{Label Sparsity Strategy}.
A large-capacity teacher network, such as CLIP trained on ImageNet-1K, produces a logit output with 1000 categories \textit{\textbf{(a)}}. Storing these complete logits as "Soft Label" in the Region Information Lookup (RIL) Table \textbf{\textit{(b)}} incurs significant storage overhead, which substantially hinders the speed of subsequent knowledge distillation for soft supervision.  To minimize storage consumption of the "Soft Label" field, we adopt two label sparsification strategies: Marginal Smoothing with Top-K (MS) \textbf{\textit{(c)}} and Marginal Re-Norm with Top-K (MR) \textit{\textbf{(d)}}, where K represents the number of most salient categories retained. MS preserves the complete information of the top-K categories while averaging the remaining probabilities. MR, in contrast, re-normalizes the probabilities of the top-K categories and sets the probabilities of the remaining categories to zero.
}

\label{fig:f5}
\centering
\end{figure*}

As depicted in Figure~\ref{fig:f4}~\textbf{(a)}, we process each input image through random cropping while recording absolute positional meta-data (coordinates, augmentation types) and soft labels in a Region Information Lookup (RIL) Table, where soft labels are obtained by feeding cropped regions into a pre-trained teacher CLIP model using the standardized prompt \textit{``A photo of a [CLASS]"} for both text and image encoders.
However, scenarios involving extensive recognition categories, such as ImageNet-1K with 1,000 classes (Figure~\ref{fig:f5}~\textbf{(a)}), suffer prohibitive I/O latency due to the excessively lengthy Soft-Label field in the RIL Table.
To mitigate this storage bottleneck, two label sparsification strategies are adopted\citep{shen2022fast,shen2023ferkd}:
\begin{enumerate}[label=\alph*.]
    \item \textbf{\textit{Marginal Smoothing with Top-K (MS)}}: as depicted in Figure~\ref{fig:f5}~\textbf{(c)}, MS strategy preserves complete information for the Top-K class categories while applying smoothing to the remaining categories:
   \begin{equation}
    \begin{aligned}
    y_c^{\text{MS}} = 
    \begin{cases}
    p_c & \text{if } c \in \{\text{Top-}K\}, \\
    1 - \frac{\sum_{c \in \{\text{Top-}K\}} p_c}{C-K} & \text{otherwise},
    \end{cases}
    \end{aligned}
    \end{equation}
   where \( p_c \) denotes the original probability for class \( c \), \( C \) is the number of class.
    \item \textbf{\textit{Marginal Re-Norm with Top-K (MR)}}: visualized in Figure~\ref{fig:f5}~\textbf{(d)}, sets the probabilities of the non-top-K categories to zero and re-normalizes the probabilities of the top-K categories:
    \begin{equation}
    \begin{aligned}
        y_c^{\text{M}} &= 
        \begin{cases}
        p_c & \text{if } c \in \{\text{Top-}K\}, \\
        0 & \text{otherwise}.
        \end{cases} \\
        y_c^{\text{MR}} &= \text{Normalize}(y_c^{\text{M}}) = \frac{y_c^{\text{M}}}{\sum_{c=1}^{C} (y_c^{\text{M}})},
    \end{aligned}
    \end{equation}
    MR ensures that the preserved probabilities sum up to 1 by explicit normalization.
\end{enumerate}
As demonstrated in Figure~\ref{fig:f5}~\textbf{(c)} and Figure~\ref{fig:f5}~\textbf{(d)}, under MS-Top-5 or MR-Top-5 setting, the storage requirement for datasets like ImageNet-1K is significantly reduced from 1000 classes to 5, allowing faster I/O during knowledge transfer.

\vspace{2mm}
\noindent
{\textbf{Advantages: Fast Iteration and Flexibility.}}
Shared Supervision Generation in this section is performed offline only once, eliminating the need for online teacher inference during KD, and the quality of the RIL Table directly impacts the effectiveness of soft supervision. 
Importantly, the RIL Table can be updated offline with more advanced configurations, such as using an ensemble of CLIP teachers, without affecting the following knowledge distillation workflow, which facilitates quick version iteration of the student CLIP model, offering a marked advantage over both native ``Prompt Learning" and existing ``Distillation-based Prompt Learning" methods.

\begin{figure}[!t]
\centering
\includegraphics[width=0.95\linewidth]{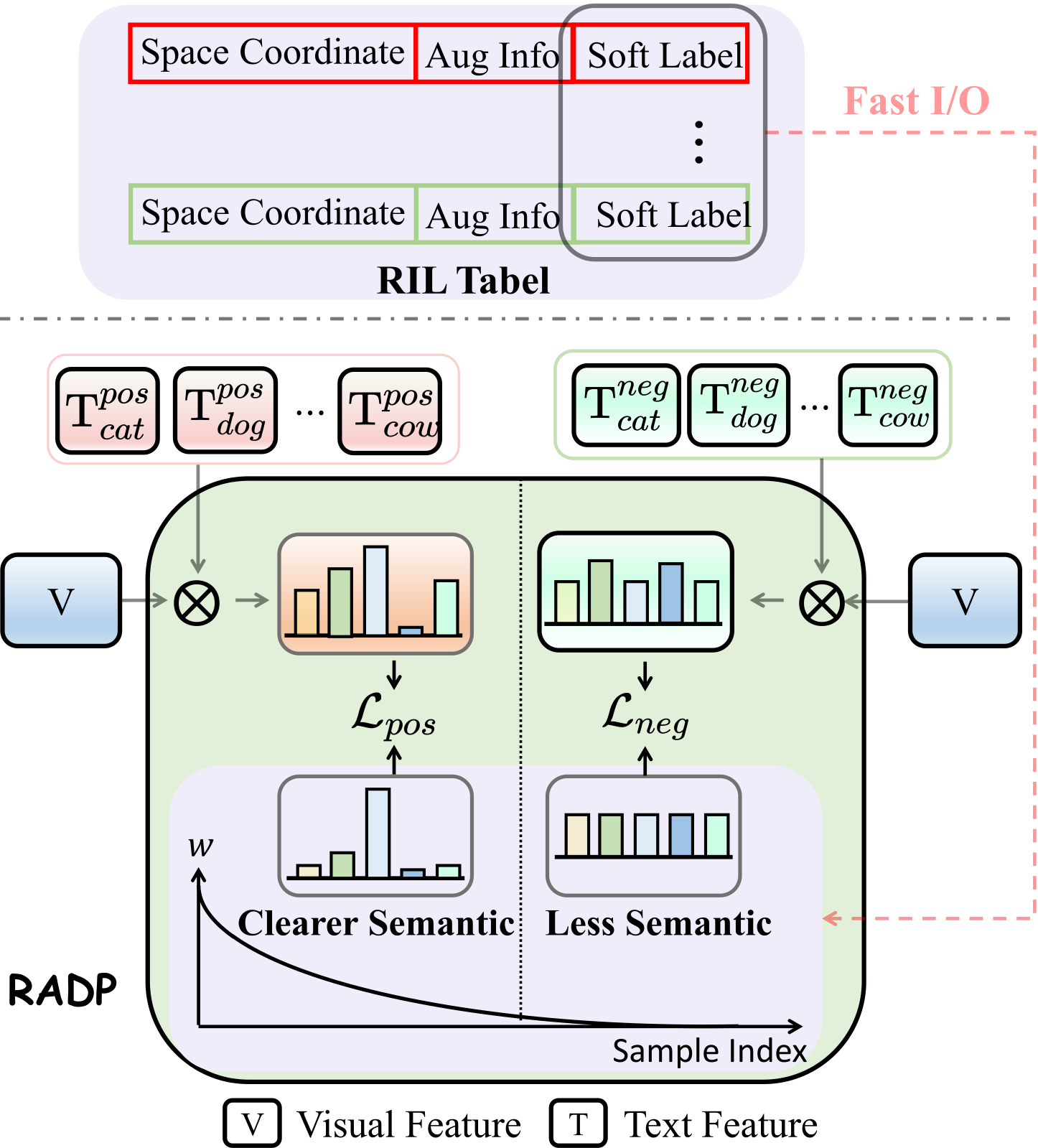}
\caption{\textbf{Region-Aware Dual-Prompt~(RADP) Learning.}
Random sampling of regional images from the Region Information Lookup (RIL) Table inevitably yields regions with varying information content, where larger weights ($w$) indicate richer information. To address this, we introduce trainable positive and negative prompts within a similarity-based training paradigm.  The positive prompts guide the student CLIP model to recognize distinct semantic categories within regions exhibiting sharp probability distributions (high information content). Conversely, the negative prompts train the student CLIP model to effectively reject regions with uniform probability distributions (low information content), teaching student CLIP to say ``No".
}
\label{fig:f6}
\end{figure}

\subsection{Region-Aware Dual-Prompt Learning}
\label{RADP}
Although Section~\ref{Shared Supervision Generation} describes the process of extracting sub-regions and offline supervision signals through image cropping, the inherent randomness of this operation combined with varying recognition difficulty across image regions may generate low-information areas containing semantically empty backgrounds. In addition to discarding such regions, we propose to exploit low-information images to train the CLIP student and recognize the "No" pattern, thereby improving its ability to reject uncertain predictions.

To achieve this, we develop a Region-Aware Dual-Prompt~(RADP) Learning that consisting of complementary positive and negative prompts, where the negative prompt specifically learns patterns associated with low-information regions. The information content of each region is quantified through entropy analysis of the teacher network's soft labels, where regions with uniform probability distributions exhibit higher entropy values, while discriminative regions yield lower entropy. To avoid binary thresholding, we construct a continuous information weight distribution:
\begin{equation}
    \begin{aligned}
        \mathbf{w} = 1 - \frac{\mathbf{H}}{\mathbf{H}_{\max}}
    \end{aligned}
    \label{weight}
\end{equation}
where \( \mathbf{w} \in [0,1] \) denotes the information weight (higher values indicate richer semantic content), \( \mathbf{H} = -\sum_{c=1}^{C} p_c \log(p_c + \epsilon) \) represents the entropy of the teacher's class probability distribution \( \{p_c\}_{c=1}^C \), \( \mathbf{H}_{\max} = \log(C) \) denotes the maximum entropy under \( C \) classes, and \( \epsilon \) ensures numerical stability.

As mentioned above, the positive and negative prompts are designed to learn from high- and low-information image regions, respectively.  We introduce a template prefix $\mathbf{V}$, formulated as ``a photo of a", for the learnable positive prompt. 
Similarly, we define the template prefix for the negative prompt as ``a photo of no", denoted as $\overline{\mathbf{V}}$. Both prompts employ the pre-trained CLIP image encoder ($f$) and text encoder ($g$) to generate their respective class text features and prediction distributions:
\begin{equation}
    \begin{aligned}
        p_i = \frac{\exp \left( \cos \left( f(\mathbf{x}_i), g(\mathbf{V}(l_i)) \right) / \tau \right) }{\sum_{l_c \in \mathcal{Y}} \exp \left( \cos \left( f(\mathbf{x}_i), g(\mathbf{V}(l_c)) \right) / \tau \right)}
    \end{aligned}
\end{equation}
and
\begin{equation}
    \begin{aligned}
    \overline{p}_i = \frac{\exp \left( \cos \left( f(\mathbf{x}_i), g(\overline{\mathbf{V}}(l_i)) \right) / \tau \right) }{\sum_{l_c \in \mathcal{Y}} \exp \left( \cos \left( f(\mathbf{x}_i), g(\overline{\mathbf{V}}(l_c)) \right) / \tau \right)}
    \end{aligned}
\end{equation}
where \( p_i \) and \( \overline{p}_i \) respectively denote the positive and negative prediction distributions for region \( \mathbf{x}_i \).

For updating both positive and negative prompts, we adopt a similarity-based training paradigm, as illustrated in Figure~\ref{fig:f6}. 
The positive prompt is updated using the teacher soft-supervision signal $t_i$, which is retrieved directly from the pre-computed Region Information Lookup (RIL) Table (detailed in Section~\ref{Shared Supervision Generation}) to  capture class-inherent semantic content through entropy-weighted KL divergence:
\begin{equation}
    \begin{aligned}
        \mathcal{L}_{pos} &= \sum_{i} w_i D_{KL}(t_i || p_i) \\ &= \sum_{i} w_i \sum_{l_c \in \mathcal{Y}} t_i(l_c) \log \frac{t_i(l_c)}{p_i(l_c)}.
    \end{aligned}
\end{equation}
Conversely, the negative prompt learns to associate low-information regions (\( 1-w_i \)) with uniform predictions by minimizing:
\begin{equation}
    \mathcal{L}_{\text{neg}} = \sum_{i} (1-w_i) \sum_{l_c \in \mathcal{Y}} \frac{1}{|\mathcal{Y}|} \log\frac{1/|\mathcal{Y}|}{\overline{p}_i(l_c)}, 
\end{equation}
effectively training the student CLIP to express uncertainty for uninformative regions.

\begin{figure}[!t]
\centering
\includegraphics[width=\linewidth]{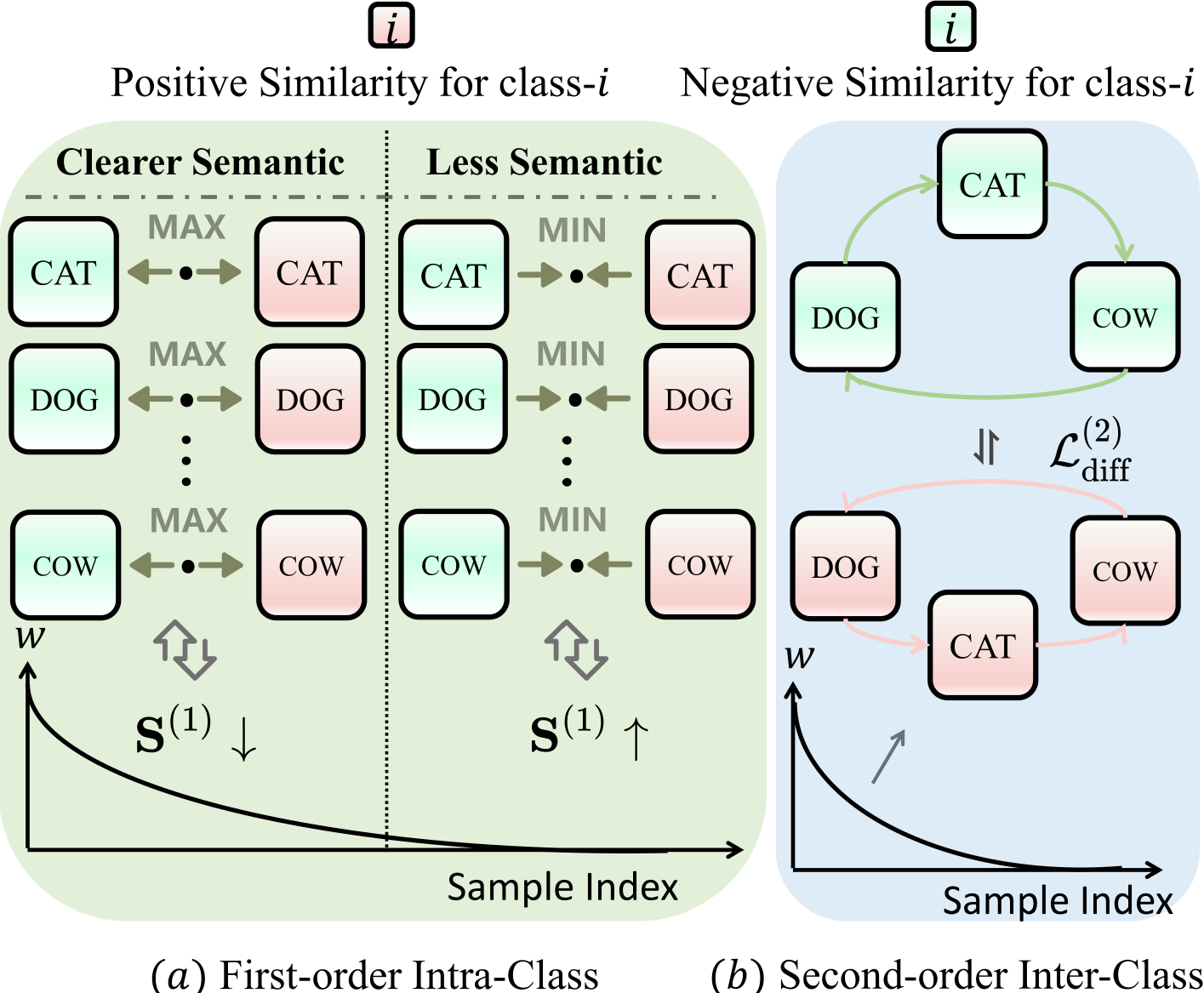}
\caption{\textbf{Prompt-Cascaded Difference~(PCD) Learning.}
\textbf{\textit{(a)}} When calculating the first-order intra-class similarity $S^{(1)}$ in the first-order difference space, we expect regions with high information content (larger values of $w$) to generate more diverse intra-class distributions between positive and negative prompt spaces (shown as \textcolor{red}{MAX} in the figure, corresponding to the $S^{(1)}$  decrease in Eq.~\ref{diff1}). 
Conversely, for regions with low information content, we expect consistent predictions between positive and negative prompt spaces (shown as \textcolor{red}{MIN} in the figure, corresponding to the $S^{(1)}$  increase in Eq.~\ref{diff1});
\textbf{\textit{(b)}} When calculating second-order intra-class similarity $S^{(2)}$ in the second-order difference space, which measures the potential inter-class relationships within the prompt space, we aim to align the spatial structures between categories across positive and negative prompt spaces in regions with high information content (larger values of $w$), which uses the inter-class inherent attributes to improve zero-shot recognition capabilities in complex scenarios.
}
\label{fig:f7}
\end{figure}

\subsection{Prompt-Cascaded Difference Learning} 
\label{PCD}
RADP follows the CLIP adaptation paradigm based on semantic similarity, which often struggle to capture fine-grained intra-class and inter-class semantic distinctions critical for complex recognition scenarios~\citep{tian2024argue}. 
To address this limitation, as demonstrated in Figure~\ref{fig:f7}, we propose \textbf{Prompt-Cascaded Difference (PCD) Learning}, a hierarchical difference space that decomposes semantic relationships through two cascaded operations: first-order intra-class difference modeling and second-order inter-class difference modeling. 

\subsubsection{First-Order Difference Modeling}
The first-order semantic difference between positive and negative prompt components is formulated as:
\begin{equation}
    \begin{aligned}
        \mathbf{D}^{(1)} = g \circ \mathbf{V} - g \circ \overline{\mathbf{V}},
    \end{aligned}
\end{equation}
where $\mathbf{V}$ and $\overline{\mathbf{V}}$ denote positive and negative prompt templates, respectively, and $g(\cdot)$ represents the student CLIP text encoder. Following $L_2$-normalization $\mathbf{D}^{(1)}_N = \mathbf{D}^{(1)}/\|\mathbf{D}^{(1)}\|_2$, the image-text similarity in this difference space is computed as:  
\begin{equation}
    \begin{aligned}
    S^{(1)}_i = \cos\left(f(x_i), \mathbf{D}^{(1)}_{N,y_1}\right),
    \end{aligned}
\end{equation}  
where $f(\cdot)$ denotes the CLIP image encoder, $y_1$ the pseudo-hard label from teacher predictions, and $\cos(\cdot)$ the cosine similarity operator. The adaptive optimization objective becomes:  
\begin{equation}
\begin{aligned}
    \mathcal{L}_{\text{diff}}^{(1)} = -\frac{1}{N}\sum_{i=1}^N \left[(1-w_i)S^{(1)}_i + \alpha w_i(1 - S^{(1)}_i)\right],
    \end{aligned}
    \label{diff1}
\end{equation}  
where $w_i \in [0,1]$ is the region information richness weight from Eq.~\ref{weight}.
As shown in Figure~\ref{fig:f7}~\textbf{(a)}, for regions with high information content ($w_i \uparrow$), we maximize intra-class diversity between positive and negative text spaces. 
In contrast, for regions with low information content ($1-w_i \uparrow$), we encourage both positive and negative prompts to produce more similar predictions toward a uniform distribution.

\subsubsection{Second-Order Difference Modeling}
We further capture inter-class differences through the second-order positive-negative space:
\begin{equation}
    \begin{aligned}
        \mathbf{D}^{(2)}_{c,c'} = \mathbf{D}^{(1)}_c - \mathbf{D}^{(1)}_{c'} \in \mathbb{R}^D, \quad \forall c \neq c'
    \end{aligned}
\end{equation}
where $\mathbf{D}^{(1)}_c$ and $\mathbf{D}^{(1)}_{c'}$ represent the first-order difference vectors for classes $c$ and $c'$, respectively, and $\mathbf{D}^{(2)}_{c,c'}$ effectively encodes the semantic direction from class $c$ to class $c'$.
After normalization $\hat{\mathbf{D}}^{(2)}_{c,c'} = \mathbf{D}^{(2)}_{c,c'}/\|\mathbf{D}^{(2)}_{c,c'}\|_2$, the structural consistency metric is defined as:  
\begin{equation}
    \begin{aligned}
        S^{(2)}_{i,c,c'} = \cos(f(x_i), \hat{\mathbf{D}}^{(2)}_{c,c'}).
    \end{aligned}
\end{equation}
where $S^{(2)}_{i,c,c'} \in \mathbb{R}$ represents the distance between the inter-class structure from category $c$ to $c'$ constructed in the positive prompt semantic space and the corresponding inter-class structure in the negative prompt semantic space.

As illustrated in Figure~\ref{fig:f7}~\textbf{(b)}, we expect consistency in inter-class structures across both positive and negative semantic spaces, which requires minimizing $S^{(2)}$. The second-order loss function $\mathcal{L}_{\text{diff}}^{(2)}$ enforces this inter-class structural consistency:
\begin{equation}
    \begin{aligned}
        \mathcal{L}_{\text{diff}}^{(2)} = \frac{1}{N(N-1)}\sum_{i=1}^N\sum_{j=1}^N & \mathbf{1}_{[y_i \neq y_j]} w_i w_j \\
        & \cdot \max(0, S^{(2)}_{i,y_i,y_j} + \delta),
    \end{aligned}
\end{equation}
where $\delta$ is a small constant.

\subsection{Optimization}
Our proposed FDBPL follows a space-for-time strategy to pre-store label information shared across multiple training epochs, alleviating the training efficiency issues in existing ``Distillation-Based Prompt Learning" CLIP adaptation paradigms. FDBPL consists of two key components: \textbf{\textit{i)}}. Region-Aware Dual-Prompt Learning, which follows a similarity-based training paradigm to independently update positive and negative prompts through $\mathcal{L}_{pos}$ and $\mathcal{L}_{neg}$, respectively, teaching CLIP to recognize inherent category semantics and to reject low-information image regions (say "No").
\textbf{\textit{ii)}}. Prompt-Cascaded Difference Learning, which follows a difference-based training paradigm to enable interaction between positive and negative prompts through $\mathcal{L}_{\text{diff}}^{(1)}$ and $\mathcal{L}_{\text{diff}}^{(2)}$, aligning intra-class diversity and inter-class consistency structures in positive and negative spaces, respectively.

Therefore, the final optimization function of our FDBPL is:
\begin{equation}
    \begin{aligned}
        \mathcal{L} = \underbrace{\mathcal{L}_{pos} + \lambda_{neg} \mathcal{L}_{neg}}_{\text{\textbf{Similarity} Training Paradigm}} + \underbrace{\lambda_{diff1} \mathcal{L}_{\text{diff}}^{(1)} + \lambda_{diff2} \mathcal{L}_{\text{diff}}^{(2)}}_{\text{\textbf{Difference} Training Paradigm}}
    \end{aligned}
    \label{loss}
\end{equation}
where $\lambda_{neg}$, $\lambda_{diff1}$ and $\lambda_{diff2}$ are the weighting factors for the negative prompt loss, the first-difference loss, and the second-difference loss, respectively.

\begin{figure*}[!b]
\centering
\includegraphics[width=\linewidth]{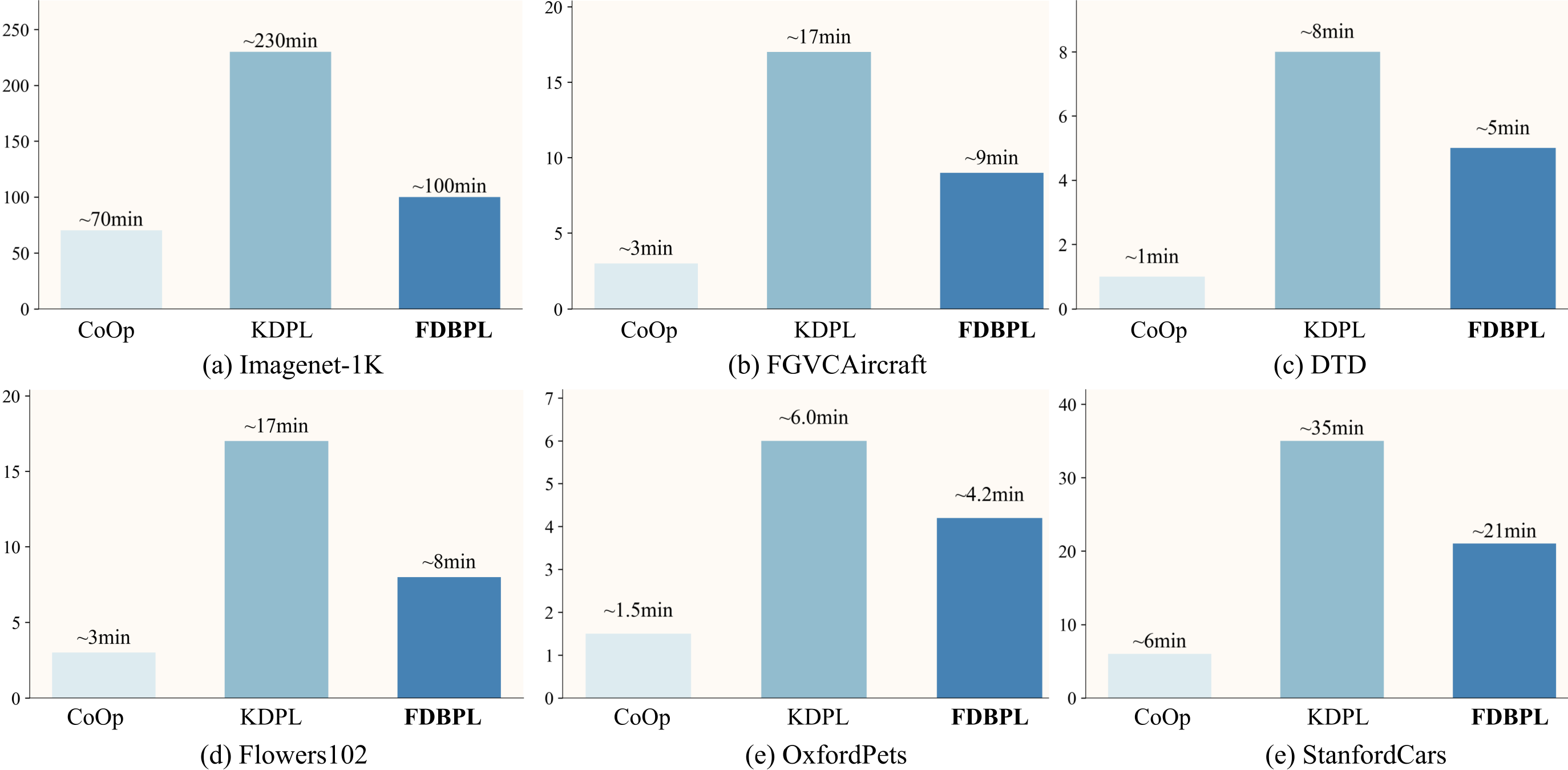}
\caption{\textbf{Training Efficiency Gain.}
Following the base-to-new protocol, this figure quantifies both the training time consumption of our proposed FDBPL and its efficiency improvement over SOTA methods. All experiments are executed on a single NVIDIA 3090 GPU, with results averaged across runs using seeds 1, 2, and 3.
}

\label{fig:f8}
\centering
\end{figure*}

\section{Experiments}
\label{Experiments}
\subsection{Experimental Setup}
\subsubsection{Datasets}
We evaluate FDBPL for zero-shot recognition on 11 datasets that include: ImageNet~\citep{deng2009imagenet} and Caltech101~\citep{fei2004learning} for general object classification; OxfordPets~\citep{parkhi2012cats}, StanfordCars~\citep{krause20133d}, Flowers102~\citep{nilsback2008automated}, Food101~\citep{bossard2014food}, and FGVCAircraft~\citep{maji2013fine} for fine-grained classification; SUN397~\citep{xiao2010sun} and UCF101~\citep{soomro2012ucf101} for action recognition; DTD~\citep{cimpoi2014describing} for texture classification; and EuroSAT~\citep{helber2019eurosat} for satellite image recognition. 
Following the few-shot learning paradigm of CoOp~\citep{zhou2022learning} and LoCoOp~\citep{miyai2023locoop}, we train FDBPL with only $N$=16 images per category.
Furthermore, we employ random cropping with $M=500$ augmentations per image to generate regional views, and a further exploration of the impact of $N$ and $M$ is presented in Section~\ref{Ablation Study}.

\subsubsection{Configuration Details}
We employ the ViT-L/14 CLIP model as the teacher network, which generates epoch-shared supervision using the prompt template ``\textit{A photo of a [class]}". The ViT-B/32 CLIP model serves as the student network. All experiments are conducted on the NVIDIA 3090 GPUs with three independent runs (random seeds {0, 1, 2}), reporting base class accuracy, novel class accuracy, and their harmonic mean (HM).
The student model is updated using the SGD optimizer, the objective function defined in Eq.~\ref{loss}, initialized with a learning rate of $10^{-1}$ following a cosine annealing schedule for 50 epochs, and the effective batch size is $128$.

\begin{figure*}[!b]
\centering
\includegraphics[width=\linewidth]{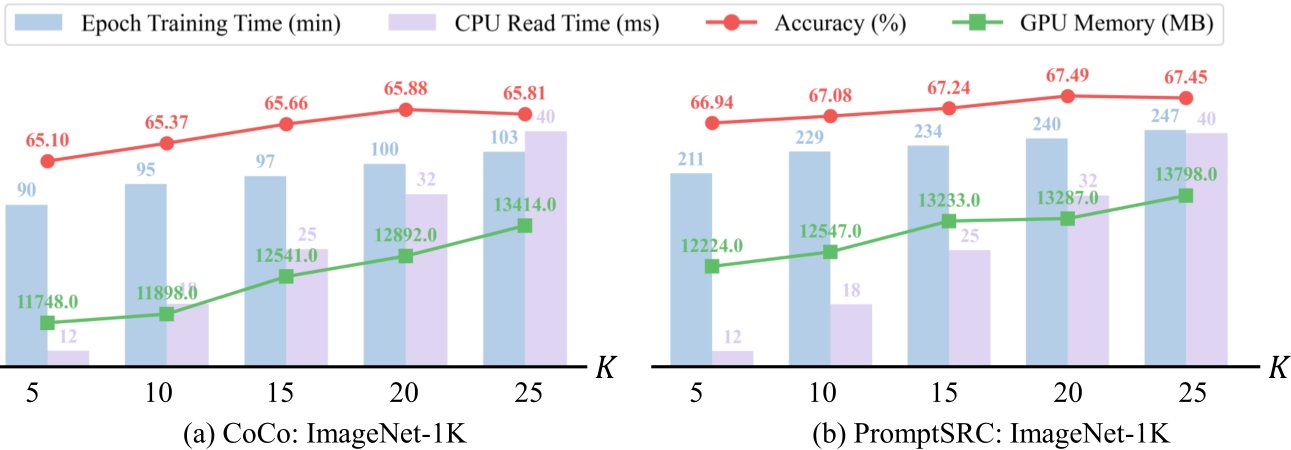}
\caption{
{
\textbf{Resource Consumption on Top-K Label Sparsification Method.}
Comprehensive resource consumption analysis under different Top-K settings ($K=5$ $\rightarrow$ $K=25$) for our Marginal Re-Norm with Top-K label sparsity strategy. The figure presents Epoch Training Time (Minutes), CPU I/O Speed (ms,  is averaged over three independent runs, each randomly reading 500 cropped offline files), GPU Memory Consumption (MB), and Accuracy (\%) on ImageNet-1K using CoCo and PromptSRC base models, illustrating the trade-off and identifying $K=20$ as optimal.
}
}
\label{fig:f9}
\end{figure*}

\subsection{Main Results}
\label{main result}
This section presents the main results of our proposed FDBPL method. We evaluate FDBPL following two established protocols: \textit{\textbf{i)}} base-to-new generalization, and \textbf{\textit{ii)}} cross-dataset evaluation.  The base-to-new protocol, adopted from CoOp~\citep{zhou2022learning} and CoCoOp~\citep{zhou2022conditional}, splits downstream task categories into non-overlapping base and new classes.  
Under the few-shot learning setting, the prompt parameters are optimized on base classes, and the performance is subsequently evaluated on both base and novel classes. The cross-dataset protocol involves optimizing the prompt parameters on ImageNet-1K and evaluating the performance on other datasets.

\begin{table*}[!b]
    \centering
    
    \caption{The \textbf{Cross-Dataset} evaluation assesses our proposed FDBPL relative to SOTA methods on various downstream datasets, where ImageNet-1K serves as the dataset for updating prompt parameters, while all other datasets are used for direct performance evaluation.
    Best results is: \firstcolor{$1_{st}$}.
    }
    
    
    \begin{subtable}{0.3\textwidth}
    \renewcommand{\arraystretch}{1.1}
        \centering
        \resizebox{\linewidth}{!}{
        \setlength{\tabcolsep}{1mm}{
        \begin{NiceTabular}{lc}
            \bottomrule[1.5px] 
            Methods & Accuracy  \\
            \midrule
            Teacher & 89.40 \\
            Student & 64.30 \\
            \midrule
            PromptSRC\citep{khattak2023self} & 65.33 \\
            PromptSRC+KDPL\citep{mistretta2024improving}$_{ECCV'24}$ & 65.10  \\
            \rowcolor{blue!6} PromptSRC+\textbf{FDBPL~(Ours)} & \firstcolor{65.85} \\
            \bottomrule[1.5px] 
        \end{NiceTabular}
        }
        }
        \caption{Flowers102}
    \end{subtable}
    \hfill
    \begin{subtable}{0.3\textwidth}
    \renewcommand{\arraystretch}{1.1}
        \centering
        \resizebox{\linewidth}{!}{
       \setlength{\tabcolsep}{1mm}{
        \begin{NiceTabular}{lc}
            \bottomrule[1.5px] 
            Methods & Accuracy  \\
            \midrule
            Teacher & 63.20  \\
            Student & 18.20  \\
            \midrule
            PromptSRC\citep{khattak2023self} & 17.17  \\
            PromptSRC+KDPL\citep{mistretta2024improving}$_{ECCV'24}$ & 18.23   \\
            \rowcolor{blue!6} PromptSRC+\textbf{FDBPL~(Ours)} & \firstcolor{18.98} \\
            \bottomrule[1.5px] 
        \end{NiceTabular}
        }
        }
        \caption{FGVCAircraft}
    \end{subtable}
    \hfill
    \begin{subtable}{0.3\textwidth}
    \renewcommand{\arraystretch}{1.1}
        \centering
        \resizebox{\linewidth}{!}{
        \setlength{\tabcolsep}{1mm}{
        \begin{NiceTabular}{lc}
            \bottomrule[1.5px] 
            Methods & Accuracy  \\
            \midrule
            Teacher & 95.60 \\
            Student & 60.40  \\
            \midrule
            PromptSRC\citep{khattak2023self} & 60.17 \\
            PromptSRC+KDPL\citep{mistretta2024improving}$_{ECCV'24}$ & 60.43 \\
            \rowcolor{blue!6} PromptSRC+\textbf{FDBPL~(Ours)} & \firstcolor{60.57} \\
            \bottomrule[1.5px] 
        \end{NiceTabular}
        }
        }
        \caption{StanfordCars}
    \end{subtable}

    \vspace{5mm}
    
    \begin{subtable}{0.3\textwidth}
    \renewcommand{\arraystretch}{1.1}
        \centering
        \resizebox{\linewidth}{!}{
        \setlength{\tabcolsep}{1mm}{
        \begin{NiceTabular}{lc}
            \bottomrule[1.5px] 
            Methods & Accuracy  \\
            \midrule
            Teacher & 93.60  \\
            Student & 79.10  \\
            \midrule
            PromptSRC\citep{khattak2023self} & 81.23  \\
            PromptSRC+KDPL\citep{mistretta2024improving}$_{ECCV'24}$ & 80.90   \\
            \rowcolor{blue!6} PromptSRC+\textbf{FDBPL~(Ours)} & \firstcolor{81.21} \\
            \bottomrule[1.5px] 
        \end{NiceTabular}
        }
        }
        \caption{Food101}
    \end{subtable}
    \hfill
    \begin{subtable}{0.3\textwidth}
    \renewcommand{\arraystretch}{1.1}
        \centering
        \resizebox{\linewidth}{!}{
        \setlength{\tabcolsep}{1mm}{
        \begin{NiceTabular}{lc}
            \bottomrule[1.5px] 
            Methods & Accuracy  \\
            \midrule
            Teacher & 76.40  \\
            Student & 62.00  \\
            \midrule
            PromptSRC\citep{khattak2023self} & 65.40 \\
            PromptSRC+KDPL\citep{mistretta2024improving}$_{ECCV'24}$ & 65.67   \\
            \rowcolor{blue!6} PromptSRC+\textbf{FDBPL~(Ours)} & \firstcolor{66.92} \\
            \bottomrule[1.5px] 
        \end{NiceTabular}
        }
        }
        \caption{SUN397}
    \end{subtable}
    \hfill
    \begin{subtable}{0.3\textwidth}
    \renewcommand{\arraystretch}{1.1}
        \centering
        \resizebox{\linewidth}{!}{
        \setlength{\tabcolsep}{1mm}{
        \begin{NiceTabular}{lc}
            \bottomrule[1.5px] 
            Methods & Accuracy  \\
            \midrule
            Teacher & 76.40 \\
            Student & 60.70 \\
            \midrule
            PromptSRC\citep{khattak2023self} & 63.63  \\
            PromptSRC+KDPL\citep{mistretta2024improving}$_{ECCV'24}$ & 63.40  \\
            \rowcolor{blue!6} PromptSRC+\textbf{FDBPL~(Ours)} & \firstcolor{64.37} \\
            \bottomrule[1.5px] 
        \end{NiceTabular}
        }
        }
        \caption{UCF101}
    \end{subtable}

    \label{tab:t1}
\end{table*}

\subsubsection{{Training Efficiency Analysis}}
\label{Training Efficiency Analysis}
As discussed in Section~\ref{Introduction}, existing distillation-based prompt learning paradigms often sacrifice the inherent training efficiency of native prompt learning due to the repeated online inference of teacher networks required for soft supervision throughout training.  Our FDBPL addresses this inefficiency through a space-for-time strategy, pre-storing shareable supervision information. Figure~\ref{fig:f8} compares training times across various downstream datasets and reveals the following:
\begin{enumerate}[label=\alph*.]
    \item Existing state-of-the-art (SOTA) distillation-based prompt learning methods (e.g., KDPL) incur longer training times compared to CoOp due to the added knowledge distillation stage.  For example, training time increases by \textbf{+69}\% and \textbf{+82}\% on ImageNet-1K and StanfordCars datasets, respectively.
    \item FDBPL significantly improves training efficiency by exploiting fast I/O and retrieving shared soft supervision information from a pre-computed Regional Image Logits (RIL) Table, thereby reducing the frequency of teacher network inference. Although Figure~\ref{fig:f8} shows substantial efficiency gains, FDBPL still exists with a slight overhead compared to CoOp.
\end{enumerate}
We further analyze the trade-off between this added overhead and zero-shot performance gains.
As detailed in Section~\ref{Shared Supervision Generation}, our label sparsity strategy employs Marginal Smoothing with Top-$K$ (MS) and Marginal Re-Norm with Top-$K$ (MR) to compress the soft label information, reducing I/O costs, where $K$ denotes the number of preserved most salient class signals. 
{Figure~\ref{fig:f9}  presents the results of this analysis, illustrating key performance metrics across a range of Top-K values (specifically, $K=$5, 10, 15, 20, and 25). The evaluated metrics include: Epoch Training Time (minutes), CPU Read Time (ms), model Accuracy (\%), and GPU Memory Consumption (MB). The CPU Read Time, in particular, is measured by averaging the duration of three independent trials, each involving the random reading of 500 cropped offline supervision files:}
\begin{enumerate}[label=\alph*.]
    \item  Increasing $K$ (from 5 to 20) progressively improves image recognition performance by incorporating more fine-grained class information, leading to more accurate soft supervision. 
    \item Excessive $K$ values (e.g., $K=25$) introduce indistinct inter-class signals that provide negligible benefits for images with limited semantic content.
    \item $K=20$ achieves optimal efficiency-performance balance, while $K=5$ offers an alternative to resource-constrained scenarios, demonstrating the operational flexibility of FDBPL.
\end{enumerate}

\begin{table*}[!t]
\caption{
Under the \textbf{Base-to-New} experimental setting, we evaluate our proposed FDBPL against SOTA methods by partitioning downstream dataset classes into seen and unseen sets. Few-shot learning updates FDBPL using only samples from the seen classes, while subsequent zero-shot recognition is evaluated on the unseen classes.
Best results is: \firstcolor{$1_{st}$}.
}

\begin{subtable}[t]{0.315\linewidth}
\centering
\renewcommand{\arraystretch}{1.1}
\resizebox{\linewidth}{!}{
\begin{NiceTabular}{lccc}
        \bottomrule[1.5px]
        {\textbf{Method}}  & {\textbf{Base}}   & \textbf{Novel} & \textbf{HM} \\
        \midrule
        Teacher & 87.02 & 86.63 & 86.25 \\ 
        Student & 64.81 & 70.05 & 67.14 \\ 
        \midrule
        \multicolumn{3}{l}{\emph{Prompt Learning}} &  \\
        CoOp~\citep{zhou2022learning} & 79.02 & 64.52 & 70.31 \\ 
         {VPT}~\citep{jia2022visual}  & 75.15 & 68.02 & 70.93 \\ 
        MaPLe~\citep{khattak2023maple} & 78.26 & 69.79 & 73.43 \\ 
         {PromptSRC}~\citep{khattak2023self} & 81.17 & 70.45 & 74.97 \\ 
         \midrule
         \multicolumn{3}{l}{\emph{Distillation-Based Prompt Learning}}  & \\
        KDPL\citep{mistretta2024improving}$_{ECCV'24}$ & 77.11 & 71.61 & 73.99 \\ 
        \rowcolor{blue!6} \textbf{FDBPL} & \firstcolor{77.35} & \firstcolor{72.45} & \firstcolor{74.57} \\ 
        \bottomrule[1.5px]
\end{NiceTabular}
}
\caption{Average}
\end{subtable}
\hfill
\begin{subtable}[t]{0.315\linewidth}
\centering
\renewcommand{\arraystretch}{1.1}
\resizebox{\linewidth}{!}{
\begin{NiceTabular}{lccc}
        \bottomrule[1.5px]
        {\textbf{Method}}  & {\textbf{Base}}   & \textbf{Novel} & \textbf{HM} \\
        \midrule
        Teacher &  86.50 & 84.10 & 85.28 \\ 
        Student &  67.40 & 64.00 & 65.66 \\ 
        \midrule
        \multicolumn{3}{l}{\emph{Prompt Learning}} &  \\
        CoOp~\citep{zhou2022learning} &  70.90 & 64.53 & 67.57  \\ 
         {VPT}~\citep{jia2022visual}  & 70.30 & 64.60 & 67.27 \\ 
        MaPLe~\citep{khattak2023maple} & 71.40 & 66.70 & 68.97  \\ 
         {PromptSRC}~\citep{khattak2023self} &   72.50 & 65.93 & 69.06 \\ 
         \midrule
         \multicolumn{3}{l}{\emph{Distillation-Based Prompt Learning}}  & \\
        KDPL\citep{mistretta2024improving}$_{ECCV'24}$ &  \firstcolor{72.70} & 66.33 & 69.38 \\ 
        \rowcolor{blue!6} \textbf{FDBPL} & 72.34  & \firstcolor{67.49} & \firstcolor{69.83} \\ 
        \bottomrule[1.5px]
\end{NiceTabular}
}
\caption{ImageNet-1K}
\end{subtable}
\hfill
\begin{subtable}[t]{0.315\linewidth}
\centering
\renewcommand{\arraystretch}{1.1}
\resizebox{\linewidth}{!}{
\begin{NiceTabular}{lccc}
        \bottomrule[1.5px]
        {\textbf{Method}}  & {\textbf{Base}}   & \textbf{Novel} & \textbf{HM} \\
        \midrule
        Teacher &  94.40 & 99.20 & 96.74 \\ 
        Student & 86.90 & 96.30 & 91.35 \\ 
        \midrule
        \multicolumn{3}{l}{\emph{Prompt Learning}} &  \\
        CoOp~\citep{zhou2022learning} &  91.90 & 93.37 & 92.64  \\ 
         {VPT}~\citep{jia2022visual}  &  94.03 & 94.83 & 94.42 \\ 
        MaPLe~\citep{khattak2023maple} &  93.07 & 96.87 & 94.93  \\ 
         PromptSRC~\citep{khattak2023self} &   93.40 & 96.30 & 94.83 \\ 
         \midrule
         \multicolumn{3}{l}{\emph{Distillation-Based Prompt Learning}}  & \\
        KDPL\citep{mistretta2024improving}$_{ECCV'24}$ & 94.37 & 96.47 & 95.40 \\ 
        \rowcolor{blue!6} \textbf{FDBPL} & \firstcolor{95.36} & \firstcolor{97.12} & \firstcolor{96.23} \\ 
        \bottomrule[1.5px]
\end{NiceTabular}
}
\caption{OxfordPets}
\end{subtable}
\vspace{3mm}
\begin{subtable}[t]{0.315\linewidth}
\centering
\renewcommand{\arraystretch}{1.1}
\resizebox{\linewidth}{!}{
\begin{NiceTabular}{lccc}
        \bottomrule[1.5px]
        {\textbf{Method}}  & {\textbf{Base}}   & \textbf{Novel} & \textbf{HM} \\
        \midrule
        Teacher &  96.60 & 85.60 & 90.77 \\ 
        Student &  68.70 & 72.30 & 70.47 \\ 
        \midrule
        \multicolumn{3}{l}{\emph{Prompt Learning}} &  \\
        CoOp~\citep{zhou2022learning} & 95.13 & 59.77 & 73.41  \\ 
         {VPT}~\citep{jia2022visual}  &  88.10 & 66.03 & 75.49 \\ 
        MaPLe~\citep{khattak2023maple} &  93.37 & 69.77 & 79.86  \\ 
         {PromptSRC}~\citep{khattak2023self} &  96.20 & 71.37 & 81.95 \\ 
         \midrule
         \multicolumn{3}{l}{\emph{Distillation-Based Prompt Learning}}  & \\
        KDPL\citep{mistretta2024improving}$_{ECCV'24}$ &  94.93 & 68.83 & 79.80 \\ 
        \rowcolor{blue!6} \textbf{FDBPL} & \firstcolor{95.88} & \firstcolor{70.36} & \firstcolor{81.16} \\ 
        \bottomrule[1.5px]
\end{NiceTabular}
}
\caption{Flowers102}
\end{subtable}
\hfill
\begin{subtable}[t]{0.315\linewidth}
\centering
\renewcommand{\arraystretch}{1.1}
\resizebox{\linewidth}{!}{
\begin{NiceTabular}{lccc}
        \bottomrule[1.5px]
        {\textbf{Method}}  & {\textbf{Base}}   & \textbf{Novel} & \textbf{HM} \\
        \midrule
        Teacher &  71.40 & 64.30 & 67.67 \\ 
        Student &   20.30 & 28.30 & 23.64 \\ 
        \midrule
        \multicolumn{3}{l}{\emph{Prompt Learning}} &  \\
        CoOp~\citep{zhou2022learning} &   30.80 & 23.03 & 26.35  \\ 
         {VPT}~\citep{jia2022visual}  &  25.23 & 29.70 & 27.28 \\ 
        MaPLe~\citep{khattak2023maple} & 23.57 & 20.43 & 21.89  \\ 
         {PromptSRC}~\citep{khattak2023self}
 &  33.77 & 23.67 & 27.85 \\ 
         \midrule
         \multicolumn{3}{l}{\emph{Distillation-Based Prompt Learning}}  & \\
        KDPL\citep{mistretta2024improving}$_{ECCV'24}$ &   30.37 & \firstcolor{28.43} & 29.39 \\ 
        \rowcolor{blue!6} \textbf{FDBPL} & \firstcolor{31.19} & 28.24 & \firstcolor{29.64} \\ 
        \bottomrule[1.5px]
\end{NiceTabular}
}
\caption{FGVCAircraft}
\end{subtable}
\hfill
\begin{subtable}[t]{0.315\linewidth}
\centering
\renewcommand{\arraystretch}{1.1}
\resizebox{\linewidth}{!}{
\begin{NiceTabular}{lccc}
        \bottomrule[1.5px]
        {\textbf{Method}}  & {\textbf{Base}}   & \textbf{Novel} & \textbf{HM} \\
        \midrule
        Teacher &   78.20 & 75.00 & 76.57 \\ 
        Student &  53.20 & 53.90 & 53.55 \\ 
        \midrule
        \multicolumn{3}{l}{\emph{Prompt Learning}} &  \\
        CoOp~\citep{zhou2022learning} & 77.17 & 45.57 & 57.31  \\ 
         {VPT}~\citep{jia2022visual}  &  72.93 & 50.17 & 59.45 \\ 
        MaPLe~\citep{khattak2023maple} &   77.27 & 53.50 & 63.23  \\ 
         {PromptSRC}~\citep{khattak2023self} &   80.57 & 54.43 & 64.97 \\ 
         \midrule
         \multicolumn{3}{l}{\emph{Distillation-Based Prompt Learning}}  & \\
        KDPL\citep{mistretta2024improving}$_{ECCV'24}$ &  \firstcolor{71.93} & 50.87 & 59.59 \\ 
        \rowcolor{blue!6} \textbf{FDBPL} & 71.08 & \firstcolor{52.14} & \firstcolor{60.14} \\ 
        \bottomrule[1.5px]
\end{NiceTabular}
}
\caption{DTD}
\end{subtable}
\vspace{3mm}
\begin{subtable}[t]{0.315\linewidth}
\centering
\renewcommand{\arraystretch}{1.1}
\resizebox{\linewidth}{!}{
\begin{NiceTabular}{lccc}
        \bottomrule[1.5px]
        {\textbf{Method}}  & {\textbf{Base}}   & \textbf{Novel} & \textbf{HM} \\
        \midrule
        Teacher & 70.70  & 82.60  & 76.21 \\ 
        Student &   43.40  & 61.50  & 50.89 \\ 
        \midrule
        \multicolumn{3}{l}{\emph{Prompt Learning}} &  \\
        CoOp~\citep{zhou2022learning} & 88.37  & 49.13  & 63.15  \\ 
         {VPT}~\citep{jia2022visual}  & 76.27  & 50.03  & 60.42 \\ 
        MaPLe~\citep{khattak2023maple} & 91.17  & 66.70  & 77.04 \\ 
         {PromptSRC}~\citep{khattak2023self} &   93.77  & 61.97  & 74.62 \\ 
         \midrule
         \multicolumn{3}{l}{\emph{Distillation-Based Prompt Learning}}  & \\
        KDPL\citep{mistretta2024improving}$_{ECCV'24}$ &  73.93  & 74.17  & 74.05 \\ 
        \rowcolor{blue!6} \textbf{FDBPL} & \firstcolor{74.88} & \firstcolor{76.44} & \firstcolor{75.66} \\ 
        \bottomrule[1.5px]
\end{NiceTabular}
}
\caption{EuroSAT}
\end{subtable}
\hfill
\begin{subtable}[t]{0.315\linewidth}
\centering
\renewcommand{\arraystretch}{1.1}
\resizebox{\linewidth}{!}{
\begin{NiceTabular}{lccc}
        \bottomrule[1.5px]
        {\textbf{Method}}  & {\textbf{Base}}   & \textbf{Novel} & \textbf{HM} \\
        \midrule
        Teacher &   94.60 & 98.20 & 96.37 \\ 
        Student & 61.00 & 69.80 & 65.10 \\ 
        \midrule
        \multicolumn{3}{l}{\emph{Prompt Learning}} &  \\
        CoOp~\citep{zhou2022learning} & 71.13 & 61.73 & 66.11  \\ 
         {VPT}~\citep{jia2022visual}  & 64.07 & 69.90 & 66.86 \\ 
        MaPLe~\citep{khattak2023maple} &  67.50 & 68.03 & 67.77  \\ 
         {PromptSRC}~\citep{khattak2023self} & 72.90 & 70.00 & 71.41 \\ 
         \midrule
         \multicolumn{3}{l}{\emph{Distillation-Based Prompt Learning}}  & \\
        KDPL\citep{mistretta2024improving}$_{ECCV'24}$ &  \firstcolor{72.53} & 68.50 & \firstcolor{70.46} \\ 
        \rowcolor{blue!6} \textbf{FDBPL} & 72.22 & \firstcolor{68.74} & 70.43 \\ 
        \bottomrule[1.5px]
\end{NiceTabular}
}
\caption{StanfordCars}
\end{subtable}
\hfill
\begin{subtable}[t]{0.315\linewidth}
\centering
\renewcommand{\arraystretch}{1.1}
\resizebox{\linewidth}{!}{
\begin{NiceTabular}{lccc}
        \bottomrule[1.5px]
        {\textbf{Method}}  & {\textbf{Base}}   & \textbf{Novel} & \textbf{HM} \\
        \midrule
        Teacher &   95.50 & 96.30 & 95.90 \\ 
        Student &   84.50 & 85.70 & 85.08 \\ 
        \midrule
        \multicolumn{3}{l}{\emph{Prompt Learning}} &  \\
        CoOp~\citep{zhou2022learning} &  85.17 & 85.50 & 85.35  \\ 
         {VPT}~\citep{jia2022visual}  &   85.47 & 85.83 & 85.64 \\ 
        MaPLe~\citep{khattak2023maple} &  86.40 & 87.47 & 86.91 \\ 
         {PromptSRC}~\citep{khattak2023self} &  86.33 & 87.13 & 86.72 \\ 
         \midrule
         \multicolumn{3}{l}{\emph{Distillation-Based Prompt Learning}}  & \\
        KDPL\citep{mistretta2024improving}$_{ECCV'24}$ &  86.53 & \firstcolor{87.63} & 87.08 \\ 
        \rowcolor{blue!6} \textbf{FDBPL} & \firstcolor{86.79} & 87.51 & \firstcolor{87.15} \\ 
        \bottomrule[1.5px]
\end{NiceTabular}
}
\caption{Food101}
\end{subtable}
\vspace{3mm}
\begin{subtable}[t]{0.315\linewidth}
\centering
\renewcommand{\arraystretch}{1.1}
\resizebox{\linewidth}{!}{
\begin{NiceTabular}{lccc}
        \bottomrule[1.5px]
        {\textbf{Method}}  & {\textbf{Base}}   & \textbf{Novel} & \textbf{HM} \\
        \midrule
        Teacher &  82.00 & 85.10 & 83.52 \\ 
        Student & 69.80 & 73.10 & 71.41 \\ 
        \midrule
        \multicolumn{3}{l}{\emph{Prompt Learning}} &  \\
        CoOp~\citep{zhou2022learning} &  79.17 & 69.37 & 73.94  \\ 
         {VPT}~\citep{jia2022visual}  &  75.70 & 75.13 & 75.41 \\ 
        MaPLe~\citep{khattak2023maple} &  78.90 & 76.57 & 77.72  \\ 
         {PromptSRC}~\citep{khattak2023self} &   80.80 & 76.77 & 78.73 \\ 
         \midrule
         \multicolumn{3}{l}{\emph{Distillation-Based Prompt Learning}}  & \\
        KDPL\citep{mistretta2024improving}$_{ECCV'24}$ &  76.53 & 77.33 & 76.94 \\ 
        \rowcolor{blue!6} \textbf{FDBPL} & \firstcolor{76.64} & \firstcolor{77.98} & \firstcolor{77.29} \\ 
        \bottomrule[1.5px]
\end{NiceTabular}
}
\caption{SUN397}
\end{subtable}
\hfill
\begin{subtable}[t]{0.315\linewidth}
\centering
\renewcommand{\arraystretch}{1.1}
\resizebox{\linewidth}{!}{
\begin{NiceTabular}{lccc}
        \bottomrule[1.5px]
        {\textbf{Method}}  & {\textbf{Base}}   & \textbf{Novel} & \textbf{HM} \\
        \midrule
        Teacher &   99.20 & 97.30 & 98.24 \\ 
        Student &  93.70 & 94.00 & 93.85 \\ 
        \midrule
        \multicolumn{3}{l}{\emph{Prompt Learning}} &  \\
        CoOp~\citep{zhou2022learning} &  97.40 & 92.13 & 94.69  \\ 
         {VPT}~\citep{jia2022visual}  & 97.10 & 92.57 & 94.79 \\ 
        MaPLe~\citep{khattak2023maple} & 97.03 & 92.27 & 94.59  \\ 
         {PromptSRC}~\citep{khattak2023self} & 97.53 & 94.70 & 96.10 \\ 
         \midrule
         \multicolumn{3}{l}{\emph{Distillation-Based Prompt Learning}}  & \\
        KDPL\citep{mistretta2024improving}$_{ECCV'24}$ &   96.90 & 94.47 & 95.67 \\ 
        \rowcolor{blue!6} \textbf{FDBPL} & \firstcolor{97.26} & \firstcolor{95.12} & \firstcolor{96.18} \\ 
        \bottomrule[1.5px]
\end{NiceTabular}
}
\caption{Caltech101}
\end{subtable}
\hfill
\begin{subtable}[t]{0.315\linewidth}
\centering
\renewcommand{\arraystretch}{1.1}
\resizebox{\linewidth}{!}{
\begin{NiceTabular}{lccc}
        \bottomrule[1.5px]
        {\textbf{Method}}  & {\textbf{Base}}   & \textbf{Novel} & \textbf{HM} \\
        \midrule
        Teacher &   78.10 & 85.20 & 81.50 \\ 
        Student &  64.00 & 71.60 & 67.59 \\ 
        \midrule
        \multicolumn{3}{l}{\emph{Prompt Learning}} &  \\
        CoOp~\citep{zhou2022learning} &  82.13 & 65.60 & 72.94  \\ 
         {VPT}~\citep{jia2022visual}  & 77.43 & 69.47 & 73.22 \\ 
        MaPLe~\citep{khattak2023maple} &   81.23 & 69.40 & 74.86  \\ 
         {PromptSRC}~\citep{khattak2023self} &   85.07 & 72.63 & 78.39 \\ 
         \midrule
         \multicolumn{3}{l}{\emph{Distillation-Based Prompt Learning}}  & \\
        KDPL\citep{mistretta2024improving}$_{ECCV'24}$ &  \firstcolor{77.50} & 74.73 & 76.09 \\ 
        \rowcolor{blue!6} \textbf{FDBPL} & 77.24 & \firstcolor{75.67} & \firstcolor{76.44} \\ 
        \bottomrule[1.5px]
\end{NiceTabular}
}
\caption{UCF101}
\end{subtable}
\label{tab:t2}
\end{table*}

\begin{figure*}[!t]
\centering
\includegraphics[width=\linewidth]{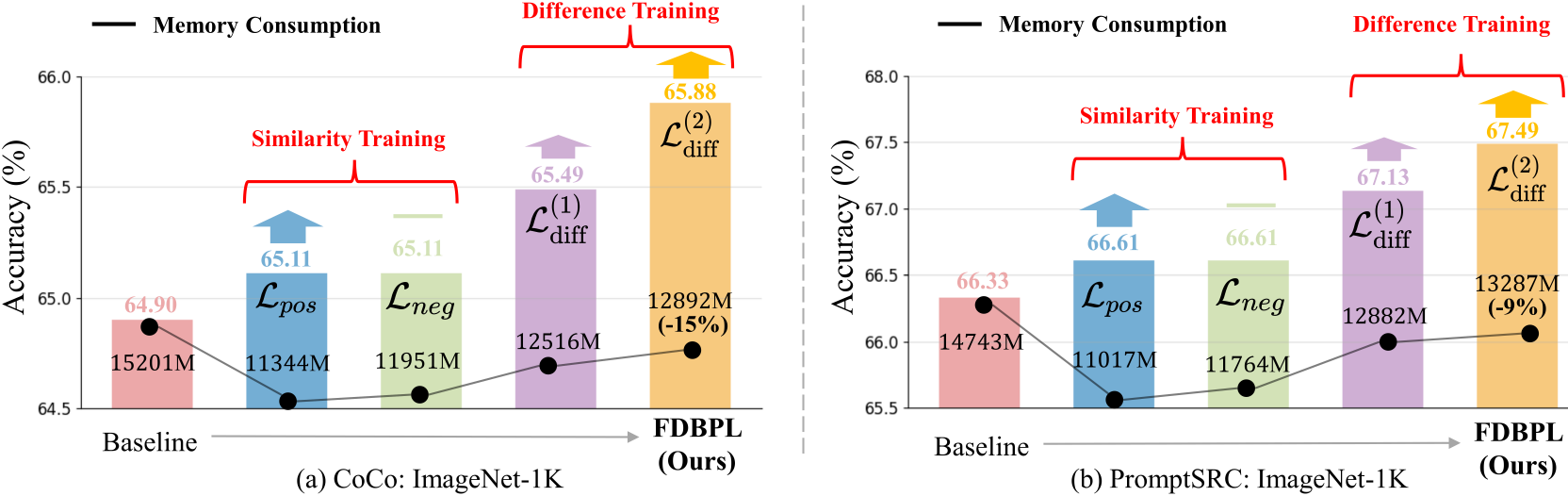}
\caption{\textbf{Ablation analysis of the FDBPL component.} 
This analysis evaluates the effectiveness of the \textbf{Similarity-Training}~($\mathcal{L}_{pos}$ and $\mathcal{L}_{neg}$) and \textbf{Difference-Training}~($\mathcal{L}_{diff}^{(1)}$ and $\mathcal{L}_{diff}^{(2)}$) strategies employed within FDBPL. Furthermore, \textbf{Memory Consumption} relative to the baseline is presented. These results collectively demonstrate the dual benefits offered by FDBPL, showcasing improvements in both zero-shot prediction performance and training efficiency.
}
\label{fig:f10}
\end{figure*}

\subsubsection{Cross-Dataset Evaluation}
In the cross-dataset evaluation setting, where FDBPL is trained on ImageNet and evaluated on other datasets, Table~\ref{tab:t1} reports comparative results between FDBPL and state-of-the-art methods on 6 different datasets. FDBPL demonstrates consistent advantages over these competing methods. The experimental results indicate that, beyond teaching CLIP to recognize correct semantic content, our FDBPL possesses unique non-semantic understanding abilities, which suggests that low-information regions likely share latent representations across datasets (e.g., background regions are dataset-independent), enabling the model to learn to reject these regions, from which FDBPL learns transferable rejection policies that benefit cross-dataset generalization.

\subsubsection{{Base-to-New Evaluation}}
Table~\ref{tab:t2} presents comparative base-to-new evaluation results across 11 datasets, yielding key observations:
\begin{enumerate}[label=\alph*.]
    \item Distillation-based methods (FDBPL and KDPL\citep{mistretta2024improving}) consistently outperform hard-supervised prompt learning approaches, verifying the effectiveness of teacher knowledge transfer for zero-shot recognition of new classes. 
    \item Compared to the existing SOTA KDPL\citep{mistretta2024improving}, FDBPL introduces low-informative regional images to teach the student CLIP model to recognize non-semantic context~(say "No"), thereby improving discriminative power for key decision regions, which yields superior performance on most novel classes. For example, on ImageNet-1K, FDBPL exceeds KDPL by \textbf{+1.13} in new class accuracy~(\%).
    \item CoOp~\citep{zhou2022learning}  achieves superior performance on base classes, likely due to their hard-label supervision leading to overfitting on the specific downstream task classes, which limits its generalization to new classes. On the other hand, FDBPL employs similarity training (RADP, Section~\ref{RADP}) and difference training (PCD, Section~\ref{PCD}) to capture inherent intra-class and inter-class structural differences, which achieves superior base class recognition performance compared to KDPL on most datasets.
\end{enumerate}

\section{Ablation Study}
\label{Ablation Study}

\subsection{Component Ablation}
The proposed FDBPL framework for region-based image prompt learning incorporates two training paradigms: similarity training (RADP, Section~\ref{RADP}) and difference training (PCD, Section~\ref{PCD}). 

Figure~\ref{fig:f10} evaluates the effectiveness and computational cost of both paradigms on ImageNet-1K dataset. 
Using KDPL as the baseline, which compromises the efficiency advantage of native prompt learning due to online teacher network inference, the proposed FDBPL efficiently stores soft labels for regional image, regaining the efficiency advantage, and reducing memory consumption.  The similarity loss, $\mathcal{L}_{pos}$, equips the student CLIP model with the ability to discern specific semantics.  $\mathcal{L}_{neg}$, trained independently of $\mathcal{L}_{pos}$, encourages higher similarity with less informative image regions, teaching the student CLIP to effectively say ``No" without impacting zero-shot performance. 
The difference training paradigm PCD employs a class-wise contrastive learning approach in the prompt space, which exploits $\mathcal{L}_{diff}^{(1)}$ to explicitly distinguish intra-class information content in a first-order difference space and $\mathcal{L}_{diff}^{(2)}$ to capture inter-class relationships in a second-order difference space, introduces additional computational overhead, and still maintains a significant memory consumption advantage compared to the baseline.

\begin{table}[!t]
\caption
{
Following the base-to-new experimental setup that employs few-shot learning principles, this ablation analysis examines how the number of training images per class affects performance.
}

\centering

\renewcommand{\arraystretch}{1.2} 

\resizebox{\linewidth}{!}{
\setlength{\tabcolsep}{1.5mm}{
\begin{NiceTabular}{l|lllll}
        \toprule[1.5px] 
        \multirow{2}{*}{{Methods}}  & \multicolumn{5}{c}{\textbf{Impact of Shots}}   \\
        \cline{2-6}
         & 1 & 2 & 4 & 8  & 16  \\
        \hline
        \hline
        \multicolumn{6}{c}{\textbf{ImageNet-1K}} \\ 
          {CoOp~\citep{zhou2022learning}} &  62.17 &  62.84 & 63.29  & 63.84  &  64.53 \\
          \rowcolor{blue!6} +\textbf{FDBPL} &  63.82$^{\mycolor[0.7]{\mathbf{+1.65}}}$ &  64.11$^{\mycolor[0.7]{\mathbf{+1.27}}}$ & 64.28$^{\mycolor[0.7]{\mathbf{+0.99}}}$  & 64.91$^{\mycolor[0.7]{\mathbf{+1.07}}}$  & \textbf{65.88}$^{\mycolor[0.7]{\mathbf{+1.35}}}$ \\
          \midrule
          \multicolumn{6}{c}{\textbf{Food-101}} \\ 
          {CoOp~\citep{zhou2022learning}} & 81.36 &  81.98 & 83.81 & 84.62 & 85.50 \\
          \rowcolor{blue!6} +\textbf{FDBPL} &  83.04$^{\mycolor[0.7]{\mathbf{+1.68}}}$ &  84.14$^{\mycolor[0.7]{\mathbf{+2.16}}}$ & 85.33$^{\mycolor[0.7]{\mathbf{+1.52}}}$  & 86.51$^{\mycolor[0.7]{\mathbf{+1.89}}}$  & \textbf{87.12}$^{\mycolor[0.7]{\mathbf{+1.62}}}$ \\
        \bottomrule[1.5px] 
    \end{NiceTabular}
}
}
\label{tab:t3}
\end{table}

\subsection{Effect of Few-Shots}
For prompt parameter updates, we follow the few-shot sampling strategy used in CoOp\citep{zhou2022learning} and CoOpOp\citep{zhou2022cocoop}, where only $N$ visible samples are exposed per class. Table~\ref{tab:t3} reports the impact of different values of $N$. We observe that as $N$ increases, both FDBPL and CoOp show performance improvements, benefiting from richer data sources. Furthermore, our FDBPL consistently outperforms CoOp across all $N$ settings, with particularly notable advantages when $N$ is small. The performance gap is attributed to the inherent differences in the training paradigm: CoOp training heavily relies on hard-label supervision, which can easily lead to overfitting and negatively impact zero-shot performance on downstream tasks. In contrast, our FDBPL draws on the knowledge with stronger generalization capabilities from the more powerful teacher CLIP model.

\begin{table}[!t]
\centering
\caption{
Ablation study on the impact of \textbf{Label Sparsification} strategies. 'Full' denotes the use of the complete soft supervision content detailed in Figure~\ref{fig:f5}~\textbf{(a)}. 
``MS" and ``MR" represent Marginal Smoothing and Marginal Re-Norm, respectively, implemented with a default $K$ value of 20. CoOp serves as the baseline method.
}
\resizebox{\linewidth}{!}{
\setlength{\tabcolsep}{3mm}{
\begin{NiceTabular}{lc|cccl} 
        \toprule[1.5px] 
        {\textbf{Methods}} & {\makecell[c]{\textbf{Hard} \\ \textbf{Sup}}}  & {\textbf{Full}} & {\textbf{MS}} & {\textbf{MR}} & {\textbf{Top-1 Acc.} (\%)} \\
        \cmidrule{1-6}
        \multicolumn{6}{c}{\textbf{SUN-397}} \\
        CoOp~\citep{zhou2022learning} & \CheckmarkBold  &    &  & & 69.37$^{\mycolor[0.7]{\mathbf{0.0}}}$ \\
        \rowcolor{blue!3} \multirow{3}{*}{\textbf{+FDBPL}}   
            & \XSolidBrush & \CheckmarkBold &   &  & 74.06$^{\mycolor[0.7]{\mathbf{+4.69}}}$ \\ 
        \rowcolor{blue!3} 
            & \XSolidBrush  &  & \CheckmarkBold &    & 73.98$^{\mycolor[0.7]{\mathbf{+4.61}}}$ \\
        \rowcolor{blue!3}  
            & \XSolidBrush  &  &   & \CheckmarkBold & {\textbf{74.21}}$^{\mycolor[0.7]{\mathbf{+4.84}}}$ \\
        \hline
        \hline
        \multicolumn{6}{c}{\textbf{DTD}} \\
             CoOp~\citep{zhou2022learning} & \CheckmarkBold  &    &  & & 45.57$^{\mycolor[0.7]{\mathbf{0.0}}}$ \\
        \rowcolor{blue!3} \multirow{3}{*}{\textbf{+FDBPL}}   
            & \XSolidBrush & \CheckmarkBold &   &  & \textbf{46.68}$^{\mycolor[0.7]{\mathbf{+1.11}}}$ \\ 
        \rowcolor{blue!3} 
            & \XSolidBrush  &  & \CheckmarkBold &    & 46.47$^{\mycolor[0.7]{\mathbf{+0.90}}}$ \\
        \rowcolor{blue!3}  
            & \XSolidBrush  &  &   & \CheckmarkBold & 46.52$^{\mycolor[0.7]{\mathbf{+0.95}}}$ \\
        \bottomrule[1.5px] 
        \end{NiceTabular}
}
}


\label{tab:t4}
\end{table}

\begin{table}[!t]
\caption
{
An ablation study evaluates the impact of varying teacher network capacities on zero-shot performance within the Base-to-new experimental framework. 
{The table further explores teacher networks with different architectures, including ViT-B/32, ViT-B/16, and ViT-L/14 models.
}
}
\centering

\renewcommand{\arraystretch}{1.1} 

\resizebox{\linewidth}{!}{
\setlength{\tabcolsep}{5mm}{
\begin{NiceTabular}{l|ccc}
        \toprule[1.5px] 
        \multirow{2}{*}{{Methods}}  & \multicolumn{3}{c}{\textbf{Teacher Network Type}}   \\
        \cline{2-4}
         & {ViT-B/32} & {ViT-B/16} & ViT-L/14 \\
        \hline
        \hline
        \multicolumn{4}{c}{\textbf{ImageNet-1K}} \\ 
          \rowcolor{blue!6} \textbf{FDBPL} &  {64.79} &  {65.12} & \textbf{65.88} \\
          \midrule
          \multicolumn{4}{c}{\textbf{Food-101}} \\ 
          \rowcolor{blue!6} \textbf{FDBPL} &  {85.98} &  {86.55} &  \textbf{87.12} \\
          \midrule
          \multicolumn{4}{c}{\textbf{SUN-397}} \\ 
          \rowcolor{blue!6} \textbf{FDBPL} &  {73.47} & {73.92} & \textbf{74.21} \\
        \bottomrule[1.5px] 
    \end{NiceTabular}
}
}
\label{tab:t5}
\end{table}

\begin{figure*}[!t]
\centering
\includegraphics[width=0.93\linewidth]{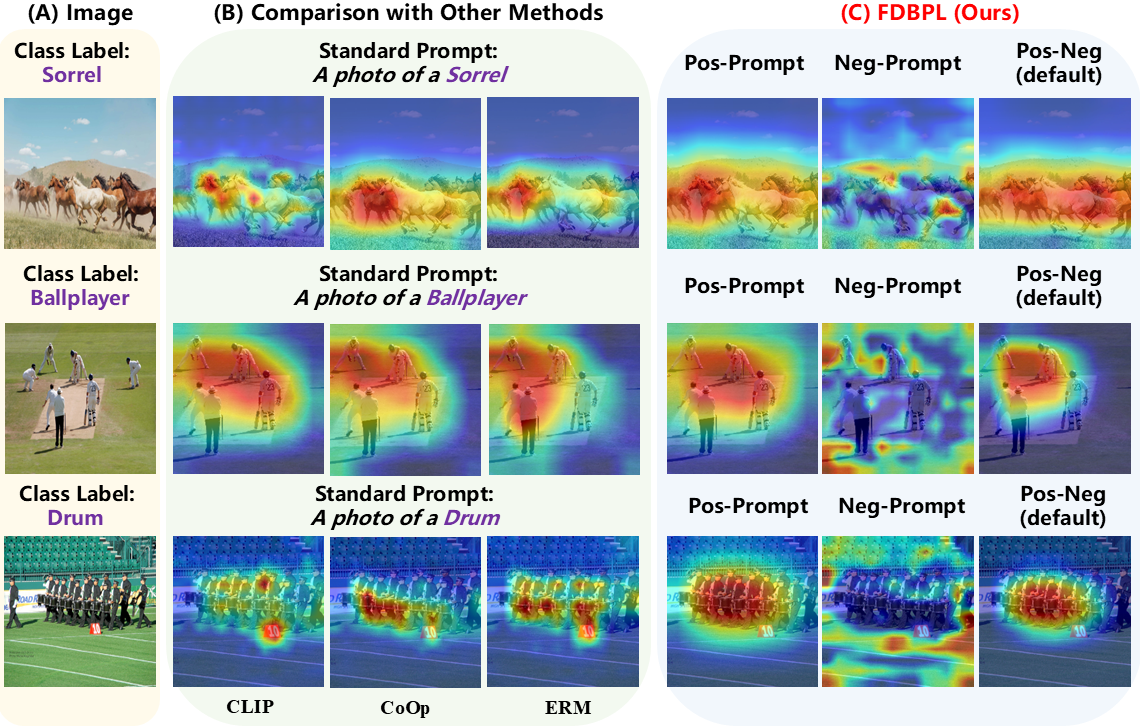}
\caption{\textbf{Attention Region Visualization}.
\textbf{\textit{(A)}}illustrates visual image inputs from ImageNet-1K and their corresponding text labels; 
\textbf{\textit{(B)}} demonstrates the attention regions focused by comparative methods. CLIP employs a hard prompt template "A photo of a [class]" to guide semantic recognition, while CoOp and ERM replace the template prefix with trainable soft tokens. We observe that these three methods either become easily distracted by background features with similar characteristics or perform incomplete recognition of target semantic content;
\textbf{\textit{(C)}} showcases the attention regions of both positive and negative prompts in our proposed FDBPL method. We focus on the auxiliary role that negative prompts play in correct identification. Specifically, negative prompts attend to non-semantic regions or teach the student CLIP to say "No," and through contrastive training with positive prompts, enable the student CLIP to pay more attention to regions that significantly impact zero-shot recognition.
}

\label{fig:f11}
\centering
\end{figure*}

\begin{table*}[!ht]
\renewcommand{\arraystretch}{1.2} 
\caption
{
Within the Base-to-new experimental setting, we conduct an ablation study to assess the contribution of the $\mathcal{L}_{\text{diff}}^{(1)}$ and $\mathcal{L}_{\text{diff}}^{(2)}$ loss functions comprising our proposed PCD learning. 
Best results is: \firstcolor{$1_{st}$}.
}

\centering

\renewcommand{\arraystretch}{1.2} 

\resizebox{0.85\linewidth}{!}{
\setlength{\tabcolsep}{4.2mm}{
\begin{NiceTabular}{l|cccccccc}
        \toprule[1.5px] 
        \multirow{2}{*}{{Methods}} & \multicolumn{2}{c}{\textbf{PCD}} & Source & \multicolumn{5}{c}{\textbf{Target}}   \\
        \cmidrule[0.5pt](rl){2-3}
        \cmidrule[0.5pt](rl){4-4}
        \cmidrule[0.5pt](rl){5-9}
        & $\mathcal{L}_{\text{diff}}^{(1)}$ & $\mathcal{L}_{\text{diff}}^{(2)}$ & ImageNet-1k & -V2 & -S & -A & -R & Average \\
        \hline
        \hline
        Student & \XSolidBrush & \XSolidBrush & 62.00 &  54.70 & 40.80 & 29.60 & 66.00 & 47.78 \\
        Teacher &  \XSolidBrush & \XSolidBrush & 82.80 &  76.60 & 71.10 & 71.10 & 91.30 & 77.53 \\
        \midrule
         \multirow{3}{*}{{CoOp\citep{zhou2022learning}+\textbf{FDBPL}}}   & \CheckmarkBold  &   & 64.21 & 56.42 & 40.95 & 30.18 & 65.47 & 48.26 \\
           &   & \CheckmarkBold  & 64.53 & 56.78 & 41.12 & 30.45 & 65.82 & 48.54 \\
          \rowcolor{blue!6} \cellcolor{white} & \CheckmarkBold  &  \CheckmarkBold & \firstcolor{65.88} & \firstcolor{57.87} & \firstcolor{41.94} & \firstcolor{31.06} & \firstcolor{66.34} & \firstcolor{49.30} \\
          \midrule
         \multirow{3}{*}{{PromptSRC\citep{khattak2023self}+\textbf{FDBPL}}}   & \CheckmarkBold  &   & 64.37 & 56.95 & 41.25 & 30.42 & 66.87 & 48.87 \\
          &   & \CheckmarkBold  & 64.82 & 57.32 & 41.48 & 30.68 & 67.25 & 49.18 \\
          \rowcolor{blue!6} \cellcolor{white} & \CheckmarkBold  &  \CheckmarkBold & \firstcolor{66.14} & \firstcolor{58.41} & \firstcolor{42.20} & \firstcolor{31.30} & \firstcolor{68.10} & \firstcolor{50.00} \\
        \bottomrule[1.5px] 
    \end{NiceTabular}
}
}
\label{tab:t6}
\end{table*}


\subsection{Ablation Experiments of Label Sparsity}
Label sparsity strategies significantly impact offline I/O requirements and the efficiency associated with accessing shared soft supervision, consequently influencing the subsequent knowledge transfer phase effectiveness. The effect of these strategies on training time is analyzed in Section~\ref{Training Efficiency Analysis} and Figure~\ref{fig:f9}. 
In this subsection, we specifically evaluate the impact of Marginal Smoothing (MS) and Marginal Re-Norm (MR), operating under a Top-K configuration (where K=20), on downstream task performance. We compare these sparse strategies against the baseline approach involving the storage of complete soft supervision vectors (denoted Full).

As reported in Table~\ref{tab:t4}, our observations indicate that for datasets featuring a large number of classes, such as SUN-397, the ``Full" strategy not only incurs substantial I/O overhead, but also fails to outperform the more efficient MR strategy. This performance gap may arise because the complete supervision vectors potentially contain excessive redundancy or even misleading signals. 
In contrast, on the DTD dataset with fewer categories, the ``Full" strategy achieves marginally better performance (less than 0.3\% improvement) over MS and MR. The above findings suggest that MS and MR offer attractive trade-offs between downstream task performance and computational efficiency.

\begin{figure*}[!ht]
\centering
\includegraphics[width=\linewidth]{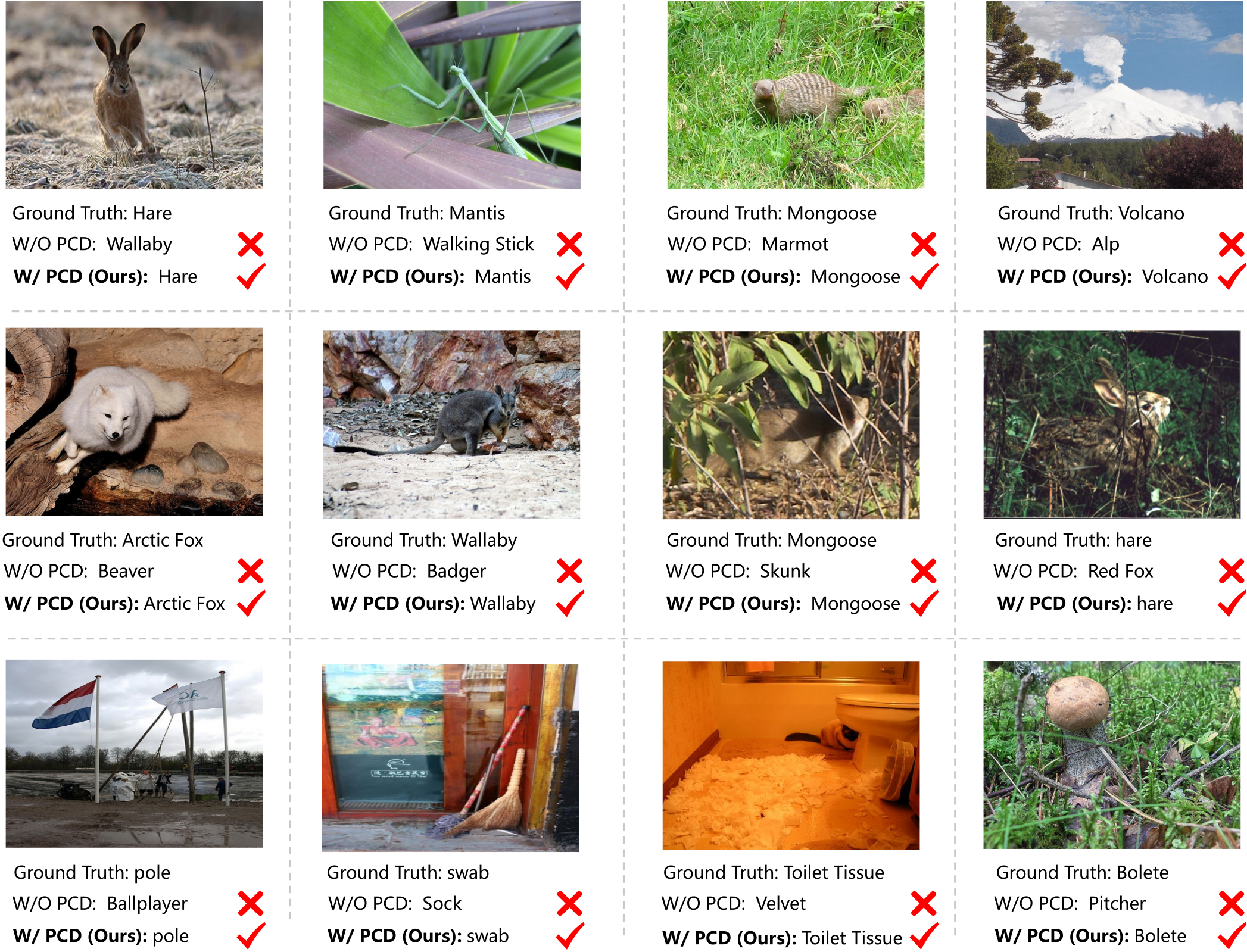}
\caption{
{
\textbf{Effect of PCD on Wrong Case Analysis.}
We present the performance of FDBPL with PCD on challenging test cases.}  
Results demonstrate that the inherent intra-class and inter-class latent relationships captured by PCD {benefit zero-shot performance in complex scenarios}.  For example, PCD improves the robustness of CLIP in distinguishing between Mantis and Walking Stick, which are difficult to differentiate when the foreground and background are highly coupled.  Similarly, PCD improve CLIP's performance in recognizing Volcano, which exhibits dynamic attributes compared to the more static Alp.
}
\label{fig:f12}
\end{figure*}

\subsection{How Important is RADP?}

\noindent
\textbf{Effect of Teacher Network Type.}
In Section~\ref{Shared Supervision Generation}, a larger CLIP teacher network (ViT-L/14) is used to generate soft supervision information for positive prompts during the shared supervision generation phase. 
{Table~\ref{tab:t5} presents an analysis of how the capacity of the teacher network (ViT-L/14, ViT-B/16, and ViT-B/32) influences the quality of the generated soft labels.}

The results on multiple downstream datasets consistently demonstrate that stronger teacher networks, particularly ViT-L/14, provide significant advantages compared to weaker alternatives such as ViT-B/32. 
{Smaller teacher networks, such as ViT-B/32 and ViT-B/16, however, enable more rapid execution of the Shared Supervision Generation process (Figure~\ref{fig:f4}~(a)). This demonstrates the flexibility of the proposed FDBPL method, which allows the selection of a teacher network based on the desired training environment and zero-shot performance requirements}.
Since both the Region Information Lookup (RIL) Table and the internal soft label fields are generated offline in a one-time process, and the RIL Table can be updated offline, the specific teacher network type does not negatively impact subsequent knowledge distillation efficiency. Based on these findings, ViT-L/14 serves as our default configuration.

\vspace{4mm}
\noindent
\textbf{Negative Prompt Behavior.}
In Figure~\ref{fig:f11}, we employ Grad-Cam to visualize the attention regions for existing prompt learning methods compared to our proposed FDBPL for given images. The observations reveal:
\begin{enumerate}[label=\alph*.]
    \item When facing inputs with complex scenes, existing methods are inevitably distracted by non-semantic content. For example,  the "horse" class and the background hills share similar color characteristics, while the person's clothing is difficult to distinguish from the grass and stands, causing the model to allocate additional computation to identify these interfering regions.
    \item Our proposed FDBPL is based on a region-specific training paradigm, where negative prompts are independently trained to emphasize low-information regions within the image. As shown in Figure~\ref{fig:f11}~\textbf{(c)}, the negative prompt successfully focuses on distracting factors that cannot be significantly differentiated using positive prompts alone. We then employ the PCD learning introduced in Section~\ref{PCD} to enable interaction between positive and negative prompt spaces (Pos-Neg), thereby encouraging the student CLIP to focus on image regions that directly influence decision-making.
\end{enumerate}

\subsection{Effect of PCD}

\noindent
\textbf{Domain Generalization.}
Domain generalization (DG) evaluates the performance of a model trained on source domains when applied to unseen target domains.  The source and target domains share the same label space, but exhibit different data distributions.  For example, target domains may contain noise or unknown corruptions. To evaluate the effectiveness of our proposed PCD learning method for DG, we conduct experiments on ImageNet-1K as the source domain and ImageNet-V2~\citep{recht2019imagenet}, ImageNet-Sketch~\citep{wang2019learning}, ImageNet-A~\citep{hendrycks2021natural}, and ImageNet-R~\citep{hendrycks2021many} as target domains.  

PCD incorporates $\mathcal{L}_{\text{diff}}^{(1)}$ and $\mathcal{L}_{\text{diff}}^{(2)}$ to capture intra-class and inter-class relationships respectively, learns prompt parameters on the source domain. 
As shown in Table~\ref{fig:f6},  the proposed FDBPL achieves superior performance when trained with a combination of $\mathcal{L}_{\text{diff}}^{(1)}$ and $\mathcal{L}_{\text{diff}}^{(2)}$, suggesting that the inherent class relationships captured PCD benefit domain generalization.

\vspace{4mm}
\noindent
{
\textbf{Wrong Case Analysis.}
Figure~\ref{fig:f12} demonstrates the effectiveness of PCD learning in recognizing complex scenes with semantically confusable elements. 
When the method without PCD that struggle to effectively distinguish between morphologically similar classes such as Hare and Wallaby, insects that blend with their environment like Mantis and Walking Stick, or dynamic attributes of Volcanoe compared to Alp. 
The proposed PCD learning mitigates these misclassifications in complex scenarios by progressively capturing potential intra-class and inter-class structures through first-order and second-order positive-negative difference spaces, which provide supplementary decision-making knowledge.
}


\section{Conclusion}
\label{sec:con}
In this paper, we propose Fast Distillation-Based Prompt Learning (FDBPL), contributing two key improvements: \textbf{\textit{i)}} FDBPL significantly accelerates training by pre-computing and storing soft supervisory contexts shared across epochs, accessed efficiently via fast I/O, thus removing the bottleneck of online teacher network inference, which also ensures flexible compatibility with existing frameworks; 
\textbf{\textit{ii)}} we introduce a positive-negative dual-prompt strategy driven by similarity and difference training, which enables effective learning from information-rich and ambiguous image regions, empowering the student CLIP model to master semantic recognition while developing the ability to reject uncertain input.
Extensive evaluations on 11 datasets, covering base-to-new generalization, cross-dataset transferability, and robustness, demonstrate the advantages of FDBPL. Our results show the effectiveness of FDBPL on unseen classes zero-shot recognition and provide an average training speed-up of $2.2\times$.

\bibliography{main}
\end{sloppypar}
\end{document}